\documentclass[accepted,specialissue]{melba}

\usepackage{mwe} 

\usepackage{amsmath,amsfonts}
\usepackage{amssymb}
\usepackage{cleveref}
\usepackage{siunitx}
\usepackage{mathtools}
\usepackage[ruled,vlined]{algorithm2e}
\usepackage{tikz}
\usepackage{hyperref}
\usepackage{url}
\usepackage{graphicx}
\usepackage{appendix}
\usepackage{booktabs}
\usepackage{amsfonts}       
\usepackage{nicefrac}
\usepackage{xcolor}         
\usepackage{algorithm2e}
\usepackage{multirow}

\newcommand{\odir}{ODIR-2019}
\newcommand{\cxr}{ChestX-ray14}
\newcommand{\bci}{BCI}
\newcommand{\clf}{$C_{f}$}
\newcommand{\clfp}{$C_{f+}$}
\newcommand{\cid}{$C_{id}$}
\newcommand{\bfx}{\mathbf{x}}

\newcommand{\bfxp}{\mathbf{x}_p}

\newcommand{\bfI}{\mathbf{I}}
\newcommand{\bft}{\mathbf{t}}

\newcommand{\bff}{\mathbf{f}}

\newcommand{\bftheta}{{\boldsymbol{\theta}}}

\newcommand{\bfs}{\mathbf{s}}

\newcommand{\mbb}[1]{\mathbb{#1}}
\newcommand{\ud}{\mathrm{d}}

\melbaid{2026:026}  
\doi{https://doi.org/10.59275/j.melba.2026-92a6}
\melbaauthors{Dombrowski and Kainz}  
\email{mischa.dombrowski@fau.de}
\volume{2026}
\firstpageno{528}  
\melbayear{2026}  
\datesubmitted{2025-12-31}  
\datepublished{2026-09-22}  

\melbaspecialissue{Medical Imaging with Deep Learning (MIDL) 2025}
\melbaspecialissueeditors{Lisa Koch, Ronald M. Summers, Chen Chen, Yan Zhuang}

\ShortHeadings{ Auditing Privacy and Fairness in Generative Medical Imaging}{Dombrowski and Kainz}

\title{A Data-Interventional Framework for Auditing Privacy and Fairness in Generative Medical Imaging}

\author{
	\firstname Mischa \surname Dombrowski\aff{1}\orcid{0000-0003-1061-8990},
	\firstname Bernhard \surname Kainz\aff{1,2}\orcid{0000-0002-7813-5023}
}
\affiliations{
	\num 1 \addr Friedrich-Alexander-Universität Erlangen-Nürnberg, Erlangen, DE \\
	\num 2 \addr Imperial College London, London, UK
}

\abstract{%
Diffusion-based synthetic data generation offers a promising route for sharing medical imaging data without releasing sensitive patient records.
However, generative models face a fundamental tension between privacy and fairness: they may memorize rare training samples, leading to privacy risks, or fail to reproduce underrepresented features, resulting in unfair synthetic distributions.
While prior work has largely focused on either memorization or fairness in isolation, their interaction remains insufficiently understood.
In this work, we introduce a data-interventional framework to systematically analyze privacy and fairness in diffusion models.
We discuss synthetic anatomical fingerprints (SAFs), rare and manually injected image features, as controlled probes to study whether models generalize sensitive attributes across identities, memorize training samples, or suppress rare signals entirely.
Across multiple conditioning modalities, we observe a consistent behavior: models either forget these fingerprints or memorize the entire image in which they appear, but do not generalize them to novel images.
To support large-scale auditing where explicit sample extraction is infeasible, we further introduce the indicator metric $t'$, which estimates a model’s susceptibility to memorization by exploiting the internal structure of the diffusion process.
By comparing conditioning signals of varying surprisal, we reveal a clear relationship between conditioning rarity and memorization behavior.
Highly surprising conditioning signals act as retrieval keys that amplify memorization, whereas low-surprisal conditioning signals systematically suppress rare features, even when these appear repeatedly in the training data.
Our findings provide actionable insights and concrete mitigation strategies for safe and fair synthetic medical data sharing.
Code is available at \url{https://github.com/MischaD/Privacy}.
}


\begin{document}

\twocolumn[\maketitle]

\section{Introduction}
Since the development of statistical models capable of representing and synthesizing new samples from existing dataset distributions~\citep{kingma2013auto,goodfellow2020generative,rombach2022high,ho2020denoising,hamamci2025generatect,guo2024maisi}, the idea of training generative models on private data and sharing \emph{only} the model or synthetic datasets has gained traction.
Such methods could address issues such as data scarcity for rare diseases, gender bias~\citep{larrazabal2020gender}, and challenges including robust domain adaptation and generalisation~\citep{wang2022metateacher}, though these benefits have not yet been demonstrated with synthetic data.
However, maintaining privacy and anonymity is essential  when working with personally identifiable information~\citep{jin2019review}.
Incorporating privacy-preserving techniques~\citep{dockhorn2022differentially} into the training process, so that only the statistical properties of private data are replicated without revealing any individual samples, therefore holds substantial promise for secure data sharing in healthcare.
However, these methods usually lead to degraded downstream performance~\citep{dockhorn2022differentially,ghalebikesabi2023differentially} and have not yet demonstrated to work at scale on high-resolution images.

Recent advances in generative modeling, including diffusion models~\citep{song2020denoising,dhariwal2021diffusion,rombach2022high,ruiz2022dreambooth}, have expanded the feasibility of sharing models directly~\citep{pinaya2022brain}.
Despite efforts of evaluating the memorization capabilities of generative models~\citep{bai2021training,dar2024unconditional,stein2024exposing}, it remains unclear to what extent shared models reproduce training samples, which would raise potential data privacy concerns.
Guarantees against privacy breaches would allow models to be trained on proprietary data and shared in place of the underlying datasets, enabling fully anonymous data sharing.
Healthcare providers could potentially distribute complex yet anonymous patient information, such as medical images, by modeling population-level data distributions.
While federated learning offers an alternative by keeping data local, it requires complex orchestration, secure aggregation, and strict coordination between institutions.
Privacy-preserving synthetic data sharing could complement federated approaches by enabling knowledge aggregation without such infrastructural overhead, though it remains an open question whether sufficient privacy guarantees can be achieved in practice~\citep{dombrowski2025enabling}.

Recent works have shown that publicly released models can inadvertently regenerate training data during sampling.
For instance, \citet{somepalli2023diffusion} and \citet{dar2024unconditional} demonstrate that diffusion models can reproduce training samples, while \citet{carlini2023extracting} illustrate how re-identifiable faces can be extracted from these datasets. 
This raises serious privacy concerns requiring robust mitigation strategies~\citep{ren2024unveiling}. 
Moreover, some generative models are explicitly designed to memorize training samples~\citep{cong2020gan}. 
Differential privacy-based strategies~\citep{dockhorn2022differentially} offer a promising avenue for mitigating these concerns. 
However, their adoption in high-resolution data synthesis remains limited due to a notable drop in distribution fidelity, restricted applicability to diffusion models, and low efficacy for multi-modal, high-resolution datasets~\citep{xie2018differentially}.
Consequently, in this work, we propose a direct approach to evaluate models, circumventing the need to obfuscate the model distribution, and thereby preserving compatibility with high-resolution, high-fidelity generative tasks.
Our approach leverages Poisson image interpolation (PII) for dataset intervention. PII is a technique commonly used in anomaly detection due to its ability to introduce highly realistic yet localized outliers into natural images~\citep{tan2021detecting,baugh2023many}.
These properties make PII particularly suitable for constructing controlled, hard-to-detect feature-level interventions that probe whether generative models memorize, suppress, or generalize rare image content.

\begin{figure*}
    \centering
    \includegraphics[width=\linewidth]{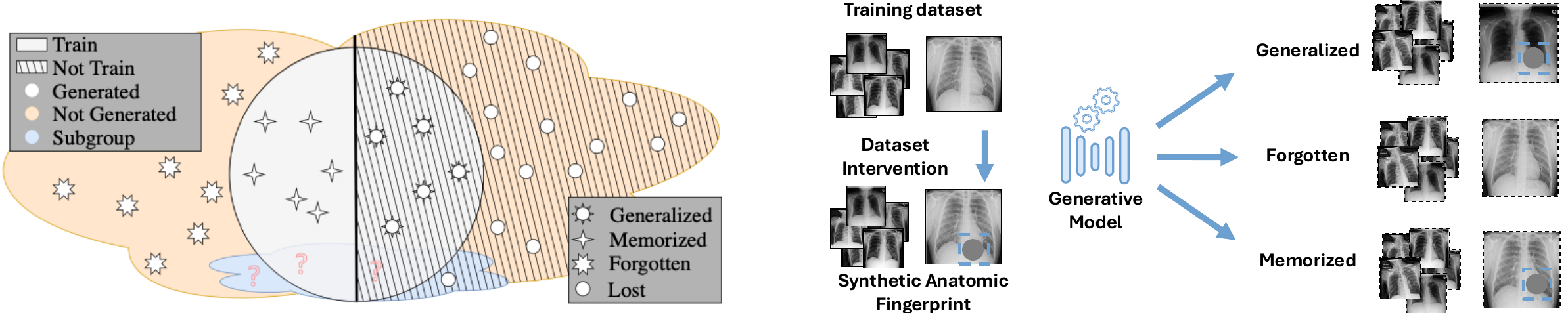}
    \caption{(Left) Training and generated images fall into one of four categories: generalized, memorized, forgotten or lost. Using SAFs, we aim to investigate the category for a selected subgroup of images.
    (Right) We use synthetic anatomical fingerprints (grey circle inside the blue rectangle) to quantify to which of these categories they belong. 
    Ideally the model learns to pick up their signal and learns to reproduce them without memorizing the entire image (generalization). 
    Memorizing the fingerprints together with their training image would lead to privacy issues. 
    Forgetting them would lead to fairness issues. 
    The dashed border indicates that images are synthetically generated.}
    \label{fig:abstract}
\end{figure*}

Not reproducing unique features has direct and largely unexplored implications for the fairness of generated data. Diffusion models trained on private datasets under strong non-memorization constraints are explicitly discouraged from reproducing rare or individual-specific characteristics. Hence, these models systematically suppress minority or outlier features, even when such features are legitimate components of the data distribution. This behavior directly conflicts with the goal of fair data generation, which requires faithful representation of both common and rare characteristics rather than their deliberate erasure.

To resolve the trade-off between privacy and fairness, models must learn to generalize. 
To approach this problem, we can conceptually divide the set of all images into two distinct categories: training set members and non-members.
After generating a dataset using a diffusion model, the images fall into one of four categories:
\begin{itemize}
\item \textbf{Lost images}: not in the training set, unavailable, potentially affecting downstream tasks but posing no privacy or fairness concerns.
\item \textbf{Memorized images}: reproduced from the private training set, raising privacy issues that require safeguards.
\item \textbf{Forgotten images}: training images not generated, potentially causing fairness issues if certain subgroups are omitted.
\item \textbf{Generalized samples}: the ideal case, where the model generates data reflecting the underlying distribution.
\end{itemize}
\noindent We illustrate this in Fig. \ref{fig:abstract}. 
To probe these categories in a controlled and observable manner, we discuss synthetic anatomical fingerprints (SAFs): rare, manually injected image features that appear only in a small number of training samples. 
They are not meant to serve as real identifiers, but rather as controllable proxies for investigating memorization.
SAFs act as surrogates for sensitive or long-tail characteristics, allowing us to test whether a generative model memorizes them together with the original image, forgets them entirely, or generalizes them across identities. 
This intervention enables a unified analysis of privacy and fairness within a single experimental framework.
For large datasets and unconditional generation, however, explicit sampling becomes computationally infeasible, and the absence of observed memorization is not sufficient evidence that memorization does not occur.
To address this, we introduce $t'$, which serves as a surrogate for the likelihood of generating fingerprint-containing samples when direct sampling is impractical.

This article extends our earlier MIDL 2025 paper, in which we introduced synthetic fingerprints as a tool to study whether generative models reproduce sensitive training data~\citep{dombrowski2025can}. 
In that work, we formulated a realistic privacy and fairness scenario for unconditional generative models, defined a formal approach to upper bound the probability of reproducing sensitive samples, and proposed the indicator metric t’ to measure this risk. 
The present work substantially broadens the scope of this exploration. 
We expand our analysis beyond unconditional generation and evaluate conditional settings, including class-, text-, and mask-guided models. 
We incorporate additional datasets to assess robustness across domains. 
We also provide a more extensive review of related work, with a particular focus on privacy auditing, membership inference, and interventional analyses that help contextualize our framework. 
In summary, 
\begin{itemize}
    \item We expand the methodological context through a more detailed literature review, integrating work in privacy auditing, membership inference, and interventional data analysis.
    \item We retain and refine the formal framework introduced in ~\citep{dombrowski2025can}, which quantifies the probability of reproducing sensitive content via SAFs.
    \item We broaden the original formulation by evaluating conditional generative models and examining how conditioning signals affect privacy and fairness behavior.
    \item We reveal a clear relationship between fairness and privacy in diffusion models and the surprisal of conditioning signals, leading to concrete safety recommendations and key mitigation strategies for safe data sharing.
\end{itemize}

\section{Related Works}

\label{Sec:background}
\label{Sec:related_work}
\paragraph{Diffusion Models}
Image generation models, such as latent diffusion models~\citep{rombach2022high}, model different levels of perturbation $p_{\sigma}(\tilde{\bfx}) \coloneqq \int p_{data}(\bfx)p_{\sigma}(\tilde{\bfx} \mid \bfx)\ud\bfx$ of the real data distribution using a noising function defined by $p_{\sigma}(\tilde{\bfx} \mid \bfx) \coloneqq \mathcal{N}(\tilde{\bfx}; \bfx, \sigma^2 \bfI)$.
Here, $\sigma$ defines the strength of the perturbation, split into $N$ steps $\sigma_{1}, \dots, \sigma_{N}$.
The assumption is that $p_{\sigma_1}(\tilde{\bfx} \mid \bfx) \sim p_{data}(\bfx)$ and $p_{\sigma_N}(\tilde{\bfx} \mid \bfx) \sim \mathcal{N}(\bfx; \textbf{0}, \sigma_N^2\bfI)$.
We define the optimization as a score matching objective by training a model $\bfs_{\bftheta}(\bfx, \sigma)$ to predict the score function $\nabla_\bfx \log p_{\sigma}(\bfx)$ for the noise level $\sigma \in \{\sigma_i\}_{i=1}^{N}$.
For sampling, this process can be reversed, for example, using Markov chain Monte Carlo methods following~\citet{song2019generative}.
\citet{song2020score} extended this to a continuous formulation by redefining the diffusion process as a process governed by an SDE and training a dense model to predict the score function.
The continuous formulation of the noising process, denoted by $p_t(\bfx)$ and $p_{st}(\bfx(t) \mid \bfx(s))$, characterizes the transition kernel from $\bfx(s)$ to $\bfx(t)$, where $0 \leq s < t \leq T$.
\citet{anderson1982reverse} showed that the reverse of this diffusion process is also a diffusion process.
\citet{song2020score} show that the reverse diffusion process of the SDE can be modeled as a deterministic process, as the marginal probabilities can be expressed deterministically in terms of the score function.
As a result, the problem simplifies to an ODE, which can be solved using any black-box numerical solver, such as the explicit Runge-Kutta method.
This enables exact likelihood computation, commonly used to estimate the likelihood of generating a sample, \emph{e.g.}, images~\citep{song2020score}.
In this context,  we propose $t'$, a more general indicator that extends this idea to approximate the likelihood of generating all samples that lead to privacy problems.

\paragraph{Memorization Detection and Mitigation}
Prior work shows memorization in diffusion models, specifically in the context of text-conditional diffusion~\citep{carlini2023extracting,somepalli2023diffusion,ren2024unveiling}. 
These findings show how text prompts can serve as keys that can retrieve near-perfect copies of training images, which raises privacy and copyright concerns. 
One of the key approaches for privacy auditing is membership inference attacks~\citep{tang2023membership,pang2023black,pang2023white,li2024towards}. 
They mostly work on reconstruction-loss style attacks, either with or without knowledge of the architecture itself. 
More importantly, they require access to the datasets that they want to infer membership on, which in practice often is not true. 

Another direction is to adopt differential privacy~\citep{dockhorn2022differentially,ghalebikesabi2023differentially,wang2024dp,liu2024efficient}.
These methods modify the training procedure to provide explicit privacy guarantees and a quantifiable privacy budget, but at the cost of substantially increased training complexity.
As a result, their practical applicability has so far been demonstrated mainly on small datasets~\citep{ghalebikesabi2023differentially}.
Similarly, differentially private fine-tuning~\citep{tsai2025differentially} can mitigate memorization but typically leads to a severe degradation in image quality.
In general, the impact of applying these methods is too disruptive to be feasible and is often highly specific to the model architecture.
Moreover, they require specialized training procedures that are incompatible with existing pretrained models, which we do not have access to.
For these reasons, we argue that improving the understanding and detection of memorization is currently a more practical and effective route toward mitigation.

Current work also actively explores the interplay between training hyperparameters and image memorization of diffusion models. 
\cite{bonnaire_why_2025} explore training the interplay between model size, training length, and dataset size and derive scaling rules for it. They specifically focus on the region when the generative model starts to generalize. 
\cite{wu_taking_2025} observed in a theoretical framework, that the learning rate of diffusion models is also a key factor for model memorization. 

To formalize and contextualize our approach, we borrow the definitions of \emph{extractable memorization} and \emph{discoverable memorization} from the natural language processing domain~\citep{nasr2023scalable,carlini2021extracting} and apply them to generative image models.
Given a model $\bfs$ with a generation routine Gen, an example $\bfxp$ from the training set $D$ is \emph{extractably memorized} if an adversary (without access to $D$) can construct a conditioning \textbf{c} that makes the model produce $\bfxp$ (\emph{i.e.}, Gen(\textbf{c}) $\approx \bfxp$).
We also adopt and extend the definition of \emph{discoverable memorization} from~\citet{nasr2023scalable} and~\citet{carlini2021extracting} to image models:
For a model $\bfs$ with generation routine Gen, an example $\bfxp \in D$, and a perturbation function from the generative model's training $p_{\sigma}(\tilde{\bfx} \mid \bfx)$, $\bfxp$ is \emph{discoverably memorized} if Gen($\tilde{\bfxp}$, $\sigma$) $\approx \bfxp$ with high probability over draws of $\tilde{\bfxp} \sim p_{\sigma}(\tilde{\bfx} \mid \bfxp)$.The strength of the perturbation function directly influences how discoverable the training images are.
Our proposed indicator $t'$ measures the susceptibility of models to discoverable memorization.
In terms of our SAF-based metrics, \clfp~(formally defined in Sec.~\ref{sec:memorization_indicator}) detections on exact identity matches correspond to discoverable memorization, while $t'$ estimates the capacity for extractable memorization by measuring how far the score function remains collapsed toward the training sample.
It can be compared to the privacy budget in differential privacy~\citep{dockhorn2022differentially}. However, unlike differential privacy methods, which only work on low-resolution images, our approach' post-hoc nature preserves image quality.

\paragraph{Fairness}
Fairness in AI is a well-explored yet unsolved problem.
Current directions in the literature for discriminative tasks suggest frameworks for benchmarking~\citep{jin2024fairmedfmfairnessbenchmarkingmedical} 
or reveal important design choices for training fair models.
Most of them provide guidance for designing and testing downstream models for fairness~\citep{yang2024limits}.
Generative models are mainly used to improve fairness~\citep{ktena2024generative,uwaeze2025generative}, 
but their own biases and unfairness remain underexplored.
Throughout this work, we use ``fairness'' to refer specifically to representational and distributional fairness, i.e.\ whether rare or long-tail features in the training data are preserved in generated outputs.
This is distinct from clinical downstream fairness notions such as demographic parity or equalized subgroup performance, which depend on task-specific evaluation.
It is often assumed that generative models learn the entire data distribution without further evaluation.
Current approaches mainly investigate fairness of generative models by looking at demographic statistics~\citep{lópezpérez2025generativemodelsfairstudy}.
Relatedly, a growing body of work studies long-tail generation in diffusion models, aiming to improve the fidelity and diversity of rare classes.
\cite{zhang2024long} propose calibrated diffusion mechanisms that leverage images from the head classes to better generate tail-class images.
\cite{samuel2024generating} address rare concept generation by selecting optimal noise seeds at inference time, enabling faithful synthesis of infrequent visual concepts without retraining.
\cite{hayden2025generative} further connect long-tail generation to robustness by guiding diffusion models toward epistemically uncertain regions, improving downstream generalization through targeted data synthesis.
While these methods substantially advance tail coverage and utility, they primarily evaluate success through class-conditional fidelity, diversity, or downstream performance gains.
Both approaches do not properly account for the fact that within certain subgroups there may be demographic differences as well. 
\cite{dombrowski_image_2024} for example explore how a retrieval-based metric can be used to quantize gender unfairness in text-to-image models.
This highlights an important gap between long-tail generative modeling and fairness analysis: improving coverage of rare features does not necessarily ensure fair representation, and may interact with privacy and demographic imbalance in subtle ways.

\paragraph{Interventional Probing}
Interventional analysis focuses on identifying causal effects by actively modifying the data a model learns from or is evaluated on.
Rather than relying on correlations in model behavior, it tests whether controlled changes to the data distribution lead to predictable changes in model outputs.
This type of analysis is most commonly applied in downstream settings, where generative models are used to synthesize targeted interventions \citep{yuan2022not,yin2023ttida,xia2024mitigating,liang2025diffusion}.
A central concept in this area is counterfactual image generation, which explicitly studies the causal and non-causal relationships between interventions and model predictions and is widely used to investigate spurious correlations.
\cite{melistas2024benchmarking} introduced a framework for benchmarking counterfactual image generation, focusing on diversity and fidelity of the generated interventions.
\cite{ak2019attribute} use GANs to generate image-level interventions by specifying semantic attributes and leveraging attention maps to localize the intervention to specific regions.
\citet{mao2021generative} perform model-level interventions in the feature space of a generative model to steer generations along semantically meaningful directions, improving robustness of downstream models.
\citet{weng2024fast} study interventions at the classifier level, using diffusion-based counterfactuals to detect and mitigate shortcut learning.
Unlike these approaches, our method focuses on data-interventions for training the diffusion model itself, not the downstream models.

\paragraph{Evaluation of Image Generation Models}
The evaluation of diffusion models is predominantly centered on fidelity by comparing distributional similarity, most commonly measured using the Fréchet Inception Distance (FID)~\citep{heusel2017gans}.
Sample diversity is typically assessed via precision and recall metrics~\citep{kynkaanniemi_improved_2019,rombach2022high,sauer_stylegan-xl_2022,dhariwal2021diffusion,peebles_scalable_2023}.
While widely adopted, these metrics offer limited interpretability and provide little insight into how individual training samples or rare features are represented.
\citet{KONZ2026103943} propose the Fréchet Radiomic Distance (FRD), a perceptual metric based on clinically meaningful radiomic features that better captures anatomical differences than FID.
Retrieval-based evaluation using image retrieval score (IRS)~\citep{dombrowski_image_2024} has been proposed as an alternative, enabling more fine-grained analysis of representational similarity.

Memorization in generative models is most commonly studied using loss-based criteria, which analyze discrepancies in likelihood or reconstruction error between training and non-training samples~\citep{bonnaire_why_2025}.
Other approaches frame memorization as a copy-detection problem and rely on trained Siamese networks to identify near-duplicates between generated and training images~\citep{dombrowski2025lcmemuniversalmodelrobust}.
While effective, these methods operate externally to the generative process and treat the model largely as a black box.

In contrast, our indicator t' directly exploits the generative mechanism of diffusion models.
By leveraging the internal structure of the diffusion process and its associated noise perturbations, t' provides a principled, model-aware signal of memorization risk.
This formulation enables a mathematically grounded analysis of memorization that yields insights beyond surface-level similarity metrics and is naturally aligned with the probabilistic foundations of diffusion models.

\paragraph{Privacy-preserving training}
Privacy-aware training strategies such as DP-SGD~\citep{dockhorn2022differentially} modify the optimization process to prevent memorization by construction.
While principled, these methods currently degrade image quality substantially and do not scale to high-resolution generation.
Our framework instead audits existing models post hoc, including pretrained models where retraining with differential privacy is impractical.

\section{Method}
\begin{figure}[t]
    \centering
    \includegraphics[width=0.99\linewidth]{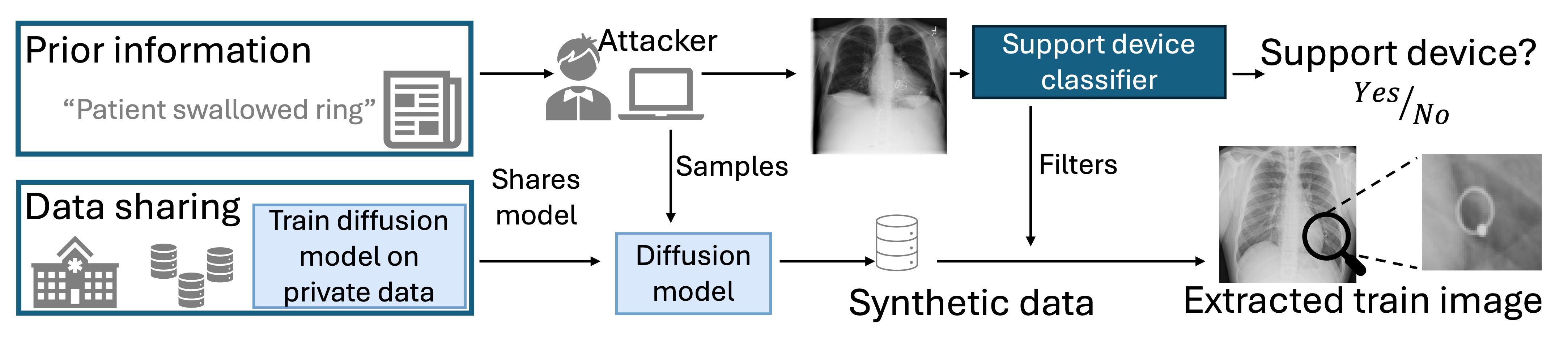}
    \caption{Extracting memorized samples from a trained diffusion model. The attacker learns that a ring is in the patient's image, uses this information, and filters generated samples until reproducing the training image.}
    \label{fig:realworldcxr}
\end{figure}

We define the key terms used throughout the paper as follows:
\begin{itemize}
    \item \textbf{Fingerprint}: An image feature that is unique to a specific person or image in the dataset (i.e., uniquely associated with a single identity, though it may appear in multiple training samples). Examples include distinctive diseases, objects, entire faces, bone structures, or medical devices. Similar to real fingerprints, their presence alone is not inherently problematic unless additional information enables identification. We use a classifier \clf~(predicting the presence of a fingerprint) and \cid~(predicting the identity) to detect these features in generated images.

    \item \textbf{Privacy}: Sharing synthetic data poses a problem if an adversary, without access to any images from the training dataset but with prior knowledge of a specific fingerprint, can extract an image from the synthetic dataset and recognize that it is memorized. Concretely, the adversary filters generated samples using knowledge of the fingerprint (e.g., knowing that a swallowed ring appears in a training image), and recognizes memorization when the generated image reproduces the fingerprint together with the appearance of the original training subject. An example of this scenario is shown in \Cref{fig:realworldcxr}.

    \item \textbf{Memorization}: Memorization refers to the pixel-wise reproduction of training images. We distinguish between full-image memorization and partial memorization. To detect partial memorization, we define fingerprints a priori and use a classifier~\clf trained to be robust against perturbations, rather than relying on pixel-wise similarity. Full-image memorization occurs when the model reproduces the entire training image.

    \item \textbf{Identity}: Case-dependent sensitive information that may be leaked if the model is shared. This can refer to the reproduction of full images, as in ChestX-ray14~\citep{wang2017chestxray} or CelebA-HQ~\citep{karras2018progressive}, but also to the identity of a person independent of image context or background.

    \item \textbf{Violation}: A privacy violation occurs when the reproduction of a fingerprint implies the identity of the original training image. Formally, this means that the presence of a fingerprint implies the presence of the associated identity.
\end{itemize}

\subsection{Conditioning Modalities.}
We consider both unconditional and conditional diffusion models. 
Unconditional image generation serves as a baseline setting, where the model is trained and sampled without any external conditioning signal. 
In addition, we evaluate diffusion models under several common conditioning paradigms. 
Specifically, we consider 
(i) class-conditioned generation, where images are generated conditioned on discrete diagnostic labels, 
(ii) text-conditioned generation, where free-form clinical reports or prompts guide synthesis, 
(iii) mask-conditioned generation, where spatial constraints such as segmentation masks are provided as input, and 
(iv) feature-conditioned generation, where images are conditioned on feature vectors extracted from pre-trained foundation models. 
These conditioning signals differ in structure and information content, but are all evaluated using the same synthetic anatomical fingerprint framework and memorization indicator \(t'\).

\subsection{Synthetic Anatomical Fingerprints.}
\begin{figure}
    \centering
    \includegraphics[width=\linewidth]{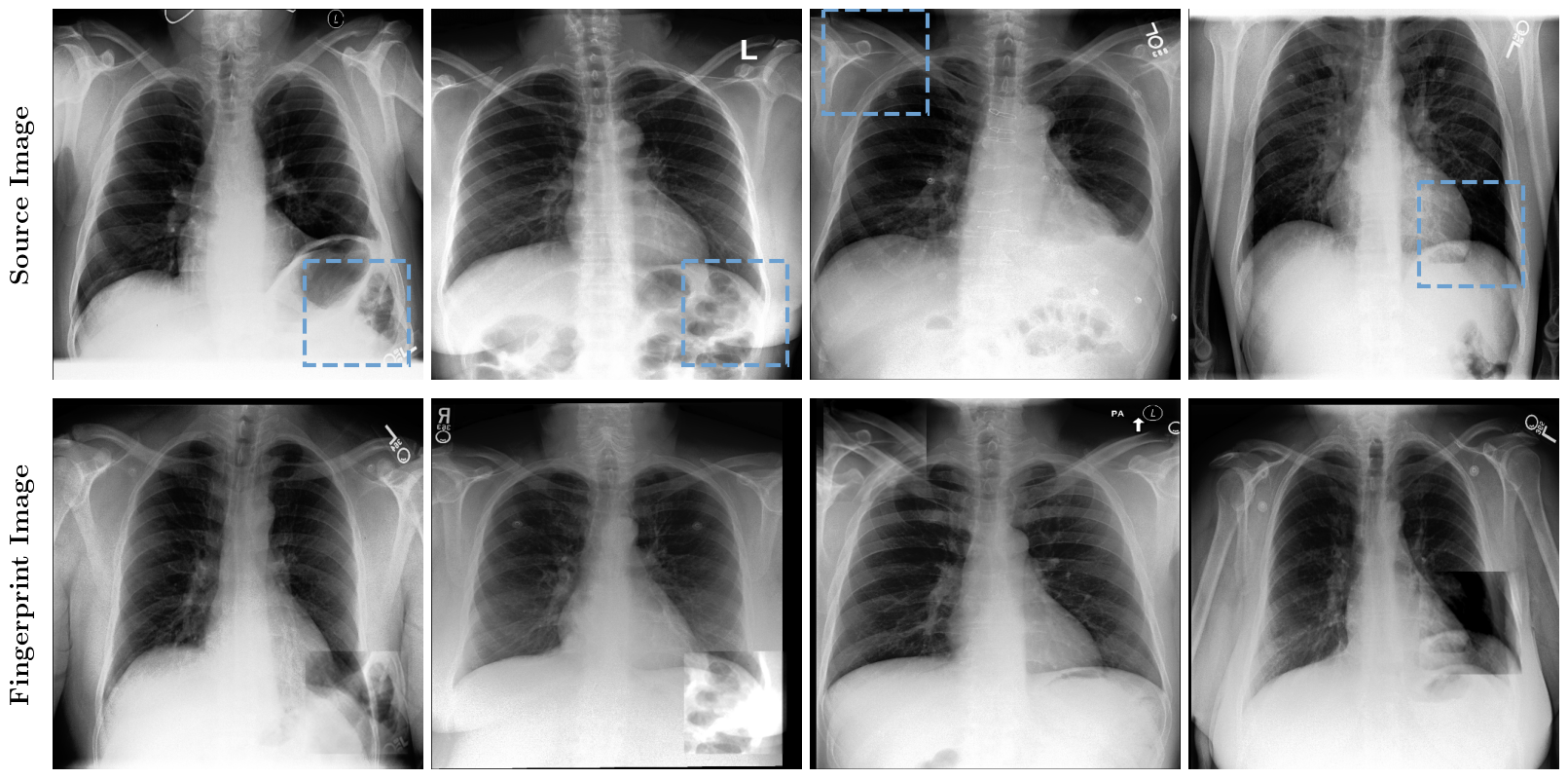}
    \caption{Example SAF: To create realistic fingerprints, we use source patches (top row) and inpaint them into target images (bottom row) using poission image editing. Depending on the similarity of source and target image, the fingerprints are barely visible.}
    \label{fig:saf_fingerprint_example}
\end{figure}

Using manually added SAFs, we study privacy and fairness within a unified experimental framework. 
Privacy concerns arise when SAFs are reproduced together with the identity of the original training image, indicating memorization.
Note that our notion of memorization includes approximate reproduction: a generated image is considered memorized if the fingerprint classifier \clf~and identity classifier \cid~both yield positive predictions, regardless of whether the reproduction is pixel-exact.
Conversely, fairness concerns emerge when SAFs are systematically absent from the generated data, suggesting that rare or unique features present in the training distribution are not retained by the model.
Our goal is to achieve generalization, where SAFs are reproduced in images with different identities.
To investigate when models start to generalize, we artificially inject detectable objects into the training data, \emph{i.e.}, SAFs.
We then train one classifier to detect these objects and another to identify the image's identity used as the target for injection.
A non-privacy-violating and fair model would reproduce the SAF on a synthetic image with a different identity than the training image containing the fingerprint.
A few examples of SAFs are shown in Fig.~\ref{fig:saf_fingerprint_example}.

To generate SAFs, we synthetically augment a single sample $\bfxp$ from the dataset $D$.
In practice, this can be any feature that appears only once in the entire training dataset, such as a ring, a deformation, or a specific medical device.
In our experiments, we consider three inpainting strategies:
(i) inserting a synthetic gray circle,
(ii) using PII to inpaint a realistic feature extracted from another image, and
(iii) incorporating known unique attributes derived from image-level metadata (e.g. only leave one female in the dataset).

\paragraph{Classifier Trainings}
\label{sec:app_training_details}
We use two binary classifiers:
(i) $C_f$, which detects the presence of a fingerprint, and
(ii) $C_{id}$, which predicts the identity associated with a training image.
Both classifiers output $1$ for a positive prediction.
We define the set of potentially memorized samples as

\begin{equation}
q := \{x : C_f(x)=1 \wedge C_{id}(x)=1\},
\label{eq:q_definition}
\end{equation}

and denote its cardinality by $|q|$.
When drawing $N_{\text{gen}}$ images from $p_s$, the expected number of such samples under a baseline assumption of uniform coverage of the training set is
\begin{equation}
    \mathbb{E}(|q|) = N_{\text{gen}} / N_{\text{train}},
\end{equation}
where $N_{\text{train}}$ is the training set size.
This baseline models a hypothetical fair model that samples each training image with equal probability; deviations between observed $|q|$ and $\mathbb{E}(|q|)$ indicate either memorization (over-representation) or suppression (under-representation) of the fingerprinted sample.
We assume that the adversary has access to the fingerprint and therefore access to $C_f$.
The attacker does not have access to $C_{id}$.
Its role is to determine whether the presence of a fingerprint implies the identity of the original training image.
This setup allows us to disentangle the memorization of the SAF from the memorization of $\bfxp$, distinguishing generalization from memorization.
To track the number of memorized samples, we define $|q|$ as the number of synthetic samples where both classifiers have a positive outcome.

\subsection{Memorization Indicator $t'$.}
\label{sec:memorization_indicator}
For large datasets and unconditional training setups, directly assessing memorization by explicitly generating samples becomes infeasible.
We therefore introduce $t’$, an indicator that captures a model’s capacity to memorize training images.
While it is possible to compute the likelihood of the exact sample (\emph{e.g.}, using numerical NLL estimation), but this does not ensure that images in the immediate neighborhood are free from privacy issues.
To address this, we propose estimating the upper bound of the likelihood of reproducing samples from the entire subspace belonging to the class of private samples.

Let $p_s(\bfxp)$ define the likelihood of the unconditional model $\bfs$ reproducing the private sample $\bfxp$ at test time.
This alone is insufficient because it does not account for slightly noisy versions of $\bfxp$, which can also pose privacy concerns.
We aim to compute $q(p)$, defined as the likelihood of reproducing any sample within $\Omega_p$.
Here, $\Omega_p$ represents the region in image space that is similar enough to $\bfxp$ to raise privacy concerns according to $q$.

In Appx. \ref{sec:estimation_method} , we show this is equivalent to:
\begin{multline}
   q(p) = \int_{\Omega_{p}} p_s(\bfx)\ud\bfx \approx \int_0^{t'} p_s(\bfx_{t,p}) \ud\bft \\ 
   \leq \sum_{i=0}^{t'} \sup_{t \in\left[t_i, t_{i+1}\right]} (\sigma_{t_{i+1}} - \sigma_{{t_i}})\mbb{E}_{p(\bfx_{t,p})}\big[p(\bfx'_{t,p})\big].
\end{multline}

To estimate $q(p)$, we observe that it depends only on the likelihood $p(\bfxp')$ and $t'$, which captures the entire region of $\Omega_p$.
$\bfxp'$ is the predicted sample of the diffusion model after applying $t$ forward diffusion steps to the private sample $\bfxp$.
This synthetic $\bfxp'$ then serves as input to the classifiers.
$\Omega_p$ is defined as the region where \cid~and \clf~both give positive predictions.
Since this region depends only on its size, $t'$ serves as an indicator of how unlikely it is to generate critical samples from the model, without the necessity to compute the exact value for $p(\bfxp')$.
Intuitively, $t'$ measures how far forward in the diffusion process we have to go for the generative model to produce a different image than the training image.

\paragraph{Assumptions and failure modes.}
The validity of $t'$ rests on the assumption that the identity classifier \cid~has learned a meaningful decision boundary for whether a generated image reproduces a specific training identity.
$t'$ can produce misleading signals in two directions:
(1)~\textbf{False sense of safety:} a low $t'$ does not guarantee the absence of memorization; it only indicates that the sampling trajectory diverges from the training sample at relatively high noise levels.
(2)~\textbf{False alarm:} if \cid~has poor specificity (\emph{e.g.}, confusing visually similar but distinct individuals), $t'$ may overestimate memorization capacity. We mitigate this through strong augmentation and the high classifier accuracies reported in Table~\ref{tab:SAFTrainingResults}.

\paragraph{Computational cost.}
Computing $t'$ for a single training sample requires $M$ reverse diffusion trajectories at each noise level (Algorithm~\ref{alg:CKB}); with $M{=}16$ this amounts to $16\times$ the cost of generating one image.
The search terminates early once the classifier no longer detects the training identity.
The main bottleneck is training the classifiers \clf~and \cid~per dataset.
Exhaustive evaluation of $t'$ over all training images is infeasible for large datasets; we recommend computing $t'$ on a targeted subset (\emph{e.g.}, rare or high-risk samples identified by domain experts) rather than exhaustively.

\paragraph{Capacity vs.\ realized memorization.}
It is important to distinguish $t'$ from \clfp.
$t'$ characterizes the learned score function around a training sample and measures the model's \emph{capacity} for memorization.
\clfp~counts the number of generated images in which \clf~detects the fingerprint, measuring \emph{realized} memorization in a finite sample.
A model can exhibit high $t'$ while producing zero \clfp~detections if the memorized region is unlikely to be reached during standard sampling.

\subsection{Class Conditional Image Generation}
\label{sec:methdo_class_cond}
Conditional diffusion models incorporate auxiliary signals such as class labels, text prompts, segmentation masks, or feature vectors to guide generation.
These signals can restrict sampling to narrow regions of the data manifold and may therefore act as keys that amplify memorization when they are rare or highly specific.

To formalize the strength and rarity of conditioning signals across modalities, we quantify their information content using concepts from information theory. 
We illustrate our idea using class conditional image generation.
For each class \(c\), we compute the class-wise entropy by evaluating the Shannon entropy of the empirical class distribution \(p(c)\), as defined in Appendix~\ref{sec:conditing}:

\begin{equation}
H(C) = - \sum_{c} p(c)\, \log_{2} p(c).
\label{eq:entropy_class}
\end{equation}

The surprisal of a specific sample with class label $c$ is then given by its information content
\begin{equation}
I(c) = -\log_{2} p(c),
\label{eq:surprisal_class}
\end{equation}
which reflects how unlikely that class is under the dataset distribution.

Similarly, we retrieve the entropy for all other considered conditioning modalities.
We present detailed derivations in Appx.~\ref{sec:conditing}.
For text-conditioned generation, we compute the entropy token-wise.
Let $t \in T$ denote a token in the vocabulary.
We estimate each token’s marginal probability $p_t$ as the fraction of prompts in the dataset that contain it:

\begin{equation}
H_{\text{text}}
= \mathbb{E}[I(X)]
= -\sum_{t \in T}
        \bigl[
            p_{t} \log_{2} p_{t}
            + (1 - p_{t}) \log_{2}(1 - p_{t})
        \bigr].
\label{eq:entropy_text}
\end{equation}

For masks, we model each pixel as an independent Bernoulli random variable with parameter $p_{uv}$.
The resulting pixel-wise entropy map is given by
\begin{equation}
H_{uv}
= -p_{uv} \log_{2}(p_{uv})
  - (1 - p_{uv}) \log_{2}(1 - p_{uv}),
\label{eq:entropy_pixel}
\end{equation}

and the total mask entropy is $\sum_{u,v} H_{uv}$.

For feature-conditioned diffusion models, we model the feature distribution as a multivariate Gaussian
(most commonly done for FID computation), compute the differential entropy, and discretize it to obtain
the discretized surprisal $I_{\text{disc}}(z_i)$ of the $i$-th feature of the conditioning vector $z$:

\begin{equation}
H_{\text{disc}}(Z)
= \frac{1}{N}\sum_{i=1}^{N} I_{\text{disc}}(z_{i}).
\label{eq:entropy_feature}
\end{equation}

\begin{figure*}[ht!]
    \centering
    \includegraphics[width=\linewidth]{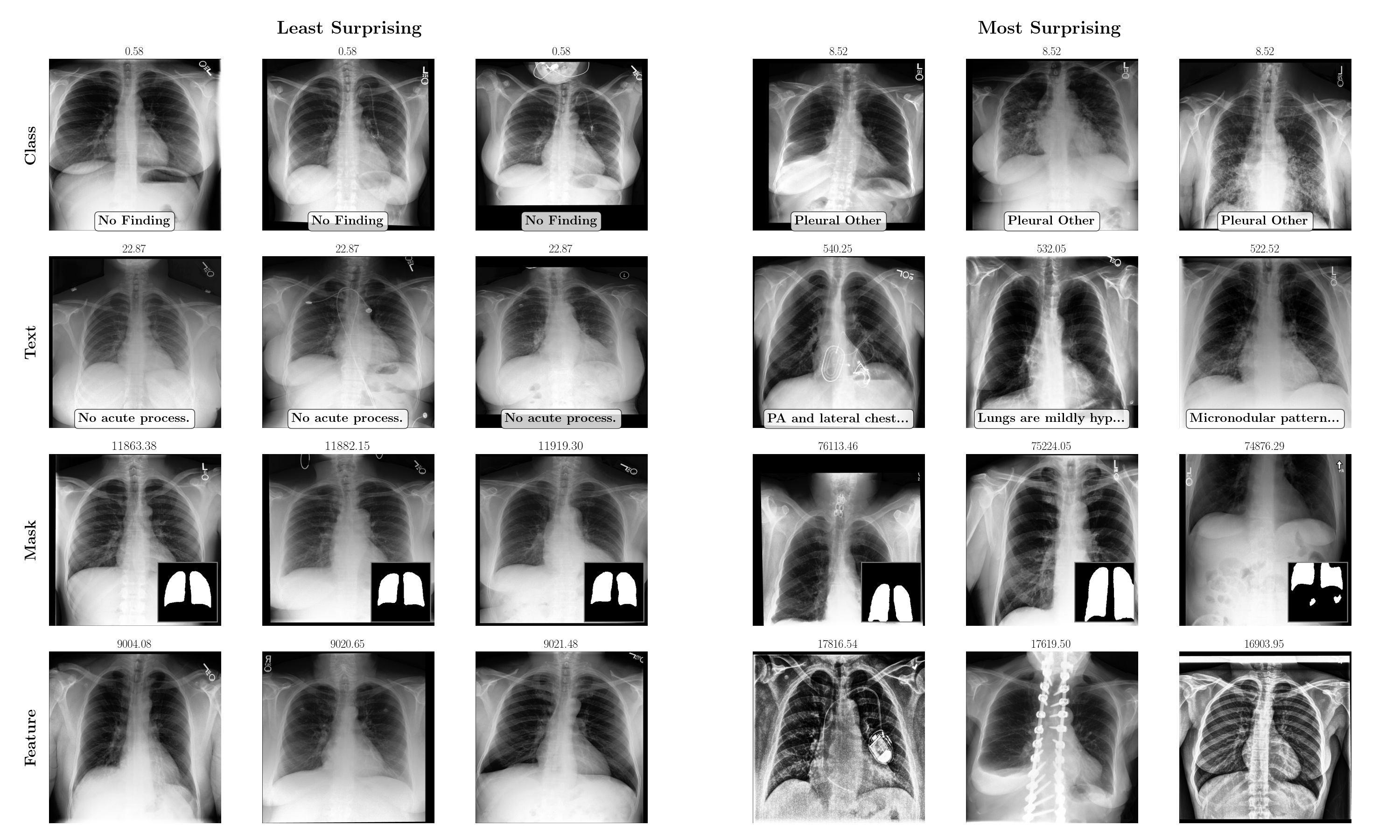}
    \caption{Example images with low (left) and high (right) surprisal.}
    \label{fig:surprisal}
\end{figure*}

Note that the absolute values of these entropies are not comparable as they all rely on different assumptions and represent different things, but their relative order is comparable. 
Specifically, we can quantize which images are the most and the least suprising for the generative model. 
We show a few examples of unsurprising and surprising conditionings in Fig. \ref{fig:surprisal}.

\label{sec:conditional}

In all conditional settings, SAFs remain the probe for memorization and fairness.
Conditioning affects how likely fingerprint-containing regions are accessed during sampling, but does not change the definition of memorization.
We therefore evaluate conditional models using the same SAF-based framework.

\section{Experiments}
\label{sec:experiments}
We structure the experimental evaluation in two stages. 
First, we analyze unconditional diffusion models to establish baseline memorization and generalization behavior under controlled variations of dataset size, training length, model capacity, and fingerprint frequency. 
Second, we extend the analysis to conditional generation, where external conditioning signals may act as keys that amplify memorization or suppress rare features. 
This design allows us to separate intrinsic model behavior from effects induced by conditioning.

Starting with unconditional image generation, we study the effects of dataset size, training duration, model capacity, and the frequency of SAFs on memorization and fairness behavior. 
For conditional generation scenarios, we include class-, text-, mask-, and feature-conditioned diffusion models. 
We specifically investigate whether the suprisal of the conditioning signal plays a critical role. 

\paragraph{Classifier Training}
The classifiers are randomly initialized ResNet50~\citep{he2016deep} architectures. 
To maximize robustness, we employ AugMix~\citep{hendrycks2019augmix}, and in the case of \cid, we inject random Gaussian noise into the training images to increase the robustness towards possible artifacts from the diffusion process. 
Furthermore, we randomly mask out patches of the same shape as the SAF by setting pixel values to zero, reducing the effect of SAF on the prediction.
Robustness is crucial for these classifiers. Even if models have a 99.9\% accuracy on the test set, they produce a not negligible amount of false predictions on a dataset with 50000 synthetic images. Therefore, we run several training sessions over different hyperparameter settings. 
Due to the simplicity of this detection and the self-supervised learning scheme, all of the classifiers trained to detect synthetic fingerprints reach an accuracy of 100\% on the test set.
Since both tasks are fairly easy binary classification tasks, we employed strong augmentation techniques to ensure that positively predicted samples from the classifiers are SAFs. 
We balanced the classification task for~\cid ~by adding SAFs to 50\% of the training images. 
For validation, we reduce this to 10\% to remain closer to the expected distribution. 
For~\cid, we chose circular masking as a training augmentation to make its predictions invariant to the presence of SAFs.

\paragraph{Classifier robustness.}
The reliability of our framework depends on the quality of \clf~and \cid.
We train multiple instances per dataset with varied hyperparameters and select the best-performing model.
As reported in Table~\ref{tab:SAFTrainingResults}, \clf~achieves 100\% test accuracy across nearly all datasets, and \cid~exceeds 99.5\% in all cases.
For the conditional experiments on real-world data (ChestX-ray14, MIMIC-CXR), we replace \cid~with the LCMem re-identification foundation model, achieving $>$96\% recall.
Despite these high accuracies, even 99.9\% accuracy produces non-negligible false predictions over 50k generated images. We therefore employ strong augmentation and, where necessary, manual verification of flagged samples.

\subsection{Unconditional Image Generation}
\label{sec:unconditional}
For our experiments on unconditional image generation, we consider the size of the training dataset, the training length, and model size as the three most impactful factors determining a model's fairness and memorization capabilities.
We first establish our method on a toy dataset, then we investigate how to tune these parameters if we want to achieve generalizability and we investigate the influence of the choice of the synthetic anatomic fingerprint. 

\paragraph{Dataset} For our initial experiments we use an a-priori selected selection of modalities from MedMNISTv2~\citep{DBLP:journals/corr/abs-2110-14795}.
For unconditional image generation, we use ChestX-ray14~\citep{wang2017chestxray}, a dataset of 112,120 frontal chest X-rays widely studied in privacy research \citep{packhauser2022deep}.
For our experiments on training length and number of fingerprints per dataset, we use three datasets with diverse modalities and sizes.
Specifically, we report results on \bci~\citep{Liu_2022_CVPR} and on \odir\footnote{\url{https://odir2019.grand-challenge.org/dataset/}}.
Data is split (60/20/20), with diffusion models sharing training data with classifiers.
For conditional image generation, we use MIMIC-CXR~\citep{johnson2019mimic}, which also comes with text impressions and precomputed lung masks~\citep{indeewara2023chest,goldberger2000physiobank}.  

\paragraph{Metrics} To evaluate generative quality, we report FID.
To quantify fairness and privacy, we compute \clfp, the number of generated images for which the SAF classifier \clf~predicts the presence of the synthetic fingerprint.
To evaluate memorization in the unconditional setting, we compute $t'$. 

\subsubsection{Toy Datasets}
\paragraph{Experimental Details} We start by training \cid,~\clf~and diffusion models on MedMNIST. 
We train an image space diffusion model based on~\citep{song2020denoising} for a fixed number of diffusion steps and until convergence. 
The diffusion component follows a compact 2D U Net design that matches the resolution of the MedMNIST images. Because the inputs are only $28\times28$, the model employs just the three outermost downsampling and upsampling blocks of the underlying U-Net. 
This reduced architecture provides sufficient capacity for the toy domains while keeping computation manageable.
A fixed training length of thirty thousand steps is used for all MedMNISTv2 datasets. Earlier experiments showed that this schedule is sufficent for the model to reliably reproduce examples from the smallest subsets, and extending training beyond this threshold offered no consistent improvement. 
Training a single model on one dataset requires roughly eleven hours on a single A100 GPU.
After convergence, each trained model is used to generate fifty thousand synthetic samples. This number of samples supports sufficient quantitative evaluation of the probability that the generative process reproduces a training example under test conditions.

\begin{table}[th]
  \caption{Training results for different MedMNIST datasets. A privacy-preserving model has a low value for $t'$, indicating a low likelihood of reproducing the sample itself, but the expected incidence of the SAF should remain close to the observed one $\mbb{E}(|q|) \sim |q|$. $|N_D|$ denotes the number of training samples.}
  \centering
\label{tab:SAFTrainingResults}
        \resizebox{\linewidth}{!}{
        \begin{tabular}{lrccccccc}
      \toprule
      \multicolumn{2}{c}{Description} & \multicolumn{2}{c}{SAF classification} & \multicolumn{5}{c}{Data synthesis} \\
       \cmidrule(r){1-2}\cmidrule(r){3-4}\cmidrule(r){5-9}
        Dataset  &$|N_D|$ & SAF (\%)& ID (\%) & $\text{FID}_{train}$ & $\text{FID}_{test}$ & $\mbb{E}(|q|)$ & $|q|$ & $t'$ \\
       \midrule
        BreastMNIST    & 546    & 100   & 98.7  & 9.2   & 62.6  & 91.6 & 57 & 0.886\\
        RetinaMNIST    & 1080   & 100   & 99.6  & 5.9   & 19.7  & 46.3 & 52 & 0.998\\
        PneumoniaMNIST & 4708   & 100   & 99.8  & 9.5   & 28.4  & 10.6 & 2  & 0.718\\
        BloodMNIST     & 11959  & 100   & 99.5  & 9.3   & 11.0  & 4.2  & 0  & 0.241\\
        OrganSMNIST    & 13940  & 99.47 & 99.8  & 19.6  & 19.7  & 3.6  & 0  & 0.582\\
        ChestMNIST     & 78468  & 99.93 & 99.8  & 3.3   & 3.9   & 0.6  & 0  & 0.206\\
     \bottomrule
  \end{tabular}%
  }
\end{table}

\paragraph{Results} The results on the toy datasets are shown in \Cref{tab:SAFTrainingResults}.
For smaller datasets, the diffusion models reproduced training images rather than generating new content. The combined prediction set \(|q| = |C_{\text{id}}^{+} \cap C_{\text{f}}^{+}|\) was non-zero only for the smallest datasets, indicating that reproduced fingerprints were always accompanied by the original training identity.
A consistent transition from memorization to generalization appears near \(|N_D| \approx 5000\). For larger datasets, \(t'\) values are small and no memorized samples are recovered.
The gap between \(\text{FID}_{train}\) and \(\text{FID}_{test}\) alone does not reliably indicate memorization: PneumoniaMNIST shows a more pronounced FID gap than RetinaMNIST yet exhibits almost no evidence of memorization.
Overall, when the number of images is limited relative to model capacity, the diffusion process may collapse onto training examples, and \(t'\) provides a practical signal for detecting this behavior.

\subsubsection{Impact of Dataset Size}
\begin{table}
        \centering
        \caption{Quantitative results on CXR data using two backbones: OD (out-of-domain) Inception and ID (in-domain) models for CXR~\citep{Cohen2022xrv}. Larger datasets reduce memorization risk, quantified by $t'$.}
        \resizebox{\linewidth}{!}{%
        \begin{tabular}{crccccccccc}
        \toprule
        &  & \multicolumn{2}{c}{Classification} & \multicolumn{2}{c}{OD (Inception)} & \multicolumn{2}{c}{ID (CXR)} & \multicolumn{3}{c}{Privacy} \\
        \cmidrule(lr){3-4}\cmidrule(lr){5-6}\cmidrule(lr){7-8}\cmidrule(lr){9-11}
         & $|N_D|$ & SAF (\%) & ID (\%) & $\text{FID}_{\text{train}}$ & $\text{FID}_{\text{test}}$ & $\text{FID}_{\text{train}}$ & $\text{FID}_{\text{test}}$ & $\mbb{E}(|q|)$ & $|q|$ & $t'$ \\
         \midrule
        \multirow{6}{*}{\rotatebox[origin=c]{90}{ChestX-ray14}} & 875  & \multirow{6}{*}{\rotatebox[origin=c]{90}{100.00}} & \multirow{6}{*}{\rotatebox[origin=c]{90}{100.00}}
                  & 15.1 & 30.3  &  1.0 &  2.3 & 34.3 & 47 & 0.75  \\
        & 1750  &&& 12.3 & 29.3  &  1.0 &  2.4 & 17.1 & 5 & 0.86  \\
        & 3500  &&& 13.6 & 32.0  &  1.2 &  2.6 & 8.6 & 1 & 0.67  \\
        & 7001  &&& 18.8 & 38.4  &  1.6 &  3.0 & 4.3 & 0 & 0.72  \\
        & 14003 &&& 22.1 & 41.4  &  1.9 &  3.3 & 2.1 & 0 & 0.66  \\
        & 28007 &&& 19.9 & 39.4  &  2.1 &  3.4 & 1.1 & 0 & 0.60  \\
            \bottomrule
        \end{tabular}%
        }
        \label{tab:datasetsizecxr}
\end{table}
    
Building on these observations from the toy datasets, we now examine how memorization behaves in a realistic large scale setting using the \cxr\ dataset. 
Here the number of available training images was varied systematically from 875 to 28,007 while keeping all other conditions fixed. 
The latent diffusion backbone was trained for 150000 steps, after which 30000 samples were generated for each configuration. 
As shown in \Cref{tab:datasetsizecxr}, dataset size directly determines the balance between memorization and generalization.
At $|N_D|=875$, the model reproduced 47 training instances with elevated $t’$ values. Increasing dataset size steadily reduced this risk; at $|N_D|=28007$ no memorized samples were detected.
A residual privacy risk remains even without detected reproductions: at $|N_D|=7001$, the high $t’$ value suggests memorization capacity persists despite zero observed SAF reproductions.
FID values are misleading in this regime, as smaller datasets yield lower FID due to memorization rather than better generative quality.
The comparison between $|q|$ and its expectation highlights a fairness concern: at $|N_D|=7001$, the expected 4.3 SAF reproductions yielded zero, indicating systematic underrepresentation of long-tail features.

\subsubsection{Impact Of Training Length}
\paragraph{Experimental Details}
To study how training duration affects synthetic image quality, we train diffusion models for multiple training lengths and compute the FID at each checkpoint. 
The model achieving the lowest FID for a given dataset is considered the best configuration and is used for all subsequent analysis.

For this initial investigation, we employ PII as the SAF. 
PII is embedded into the training data before model optimization and subsequently removed during evaluation, allowing us to quantify the impact of training length on fingerprint retention.

\begin{figure*}[t]
    \centering
    \includegraphics[width=\linewidth]{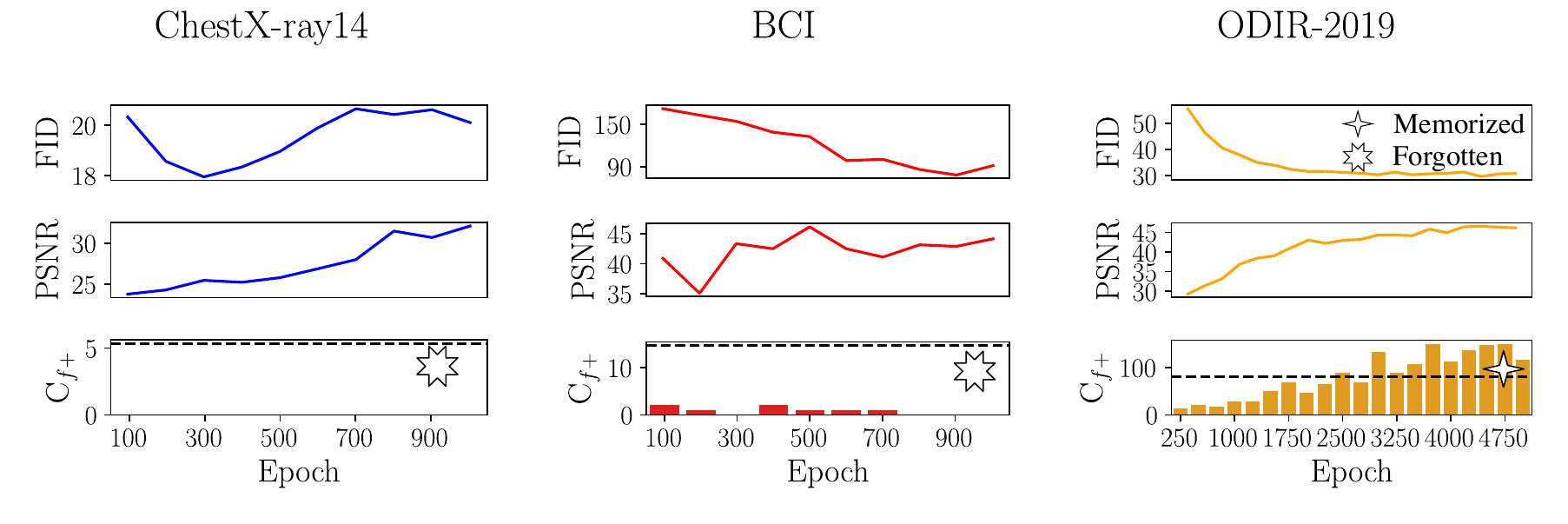}
    \caption{Impact of training length. To assess model memorization, we investigate PSNR and \clfp. The dashed line in the bottom row indicates the results of a fair model (SAFs appear equally often in training and synthetic datasets).}
    \label{fig:epochs}
\end{figure*}

\paragraph{Results}
As shown in \Cref{fig:epochs}, the effect of training length on memorization varied substantially across the three datasets.
For \cxr, extended training increased both PSNR and FID, indicating reduced sample diversity and diminished image quality. 
Despite this degradation, the model never reproduced the embedded SAF, suggesting that overtraining primarily led to mode collapse rather than direct memorization.
The \bci dataset showed a small number of positive fingerprint predictions during early epochs, but closer inspection revealed these to be false positives. 
The reduced image quality in early training likely caused spurious classifier responses. The model selected based on the lowest FID did not reproduce the SAF, and its comparatively high PSNR was driven by synthetic images containing large uniform or empty regions rather than genuine similarity to training samples.
In contrast, \odir\ demonstrated clear and consistent memorization. The model reproduced the SAF frequently, reflected by numerous positive \clfp predictions and elevated PSNR values. All detected cases corresponded to exact copies of the fingerprint-containing training sample, confirming memorization rather than incidental similarity.
Overall, the results show a dichotomy across datasets. 
Depending on the data domain and generative difficulty, diffusion models either learn to reproduce the SAF directly, introducing privacy concerns, or generate limited and low-diversity outputs that impair fairness and utility.

\subsubsection{Impact of Model Size}

\begin{table}[t]
    \centering
        \caption{
        Privacy metrics for different model sizes. Architecture details are given in \Cref{tab:model_size_architecture}.}
        \resizebox{\linewidth}{!}{%
        \begin{tabular}{lcccc}
    \toprule
    & \# Trainable parameters & $|q|$ & FID & $t'$ \\
    \midrule
    Default & \num{113675524} & 5 & 32.7 & 0.77 \\
    Model 1 & \num{77364740}  & 9 & 33.0 & 0.74 \\
    Model 2 & \num{71439108}  & 0 & 33.9 & 0.69 \\
    Model 3 & \num{49558020}  & 1 & 34.8 & 0.69 \\
    Model 4 & \num{28484612}  & 0 & 78.7 & 0.66 \\
    Model 5 & \num{28448388}  & 0 & 43.6 & 0.64 \\
    \bottomrule
    \end{tabular}%
        }
    \label{tab:model_size}
\end{table}

\paragraph{Experimental Details}
To assess how architectural capacity influences memorization independently of dataset size, we train a series of diffusion models on a fixed subset of ChestX-ray14 containing $|N_D| = 1770$ images. The experiment varies the number of trainable parameters by constructing six U-Net backbones that differ in depth, channel width, and number of down blocks. All models are trained under identical conditions and evaluated using FID, the indicator $t’$, and the number of recovered memorized samples $|q|$. 

\paragraph{Results} 
The relationship between model size, generative quality, and memorization potential is summarized in Tab. \ref{tab:model_size}, while the full architectural details are provided in the appendix.
The results reveal a clear trade-off: larger models exhibit higher $t’$ and more recovered training samples, while smaller models produce $|q|=0$ but at the cost of increasing FID.
Manual inspection confirms that smaller models avoid memorization by forgetting long-tail features rather than learning to generalize them. Reduced capacity thus protects privacy only by suppressing rare patterns.

\subsubsection{Impact of Type of Fingerprint}
\begin{table}
\caption{Memorization results for all three datasets. Results are averaged over three different runs. All \cxr\ and \bci\ runs result in fairness issues due to the complete lack of reproducing the SAFs.}
\resizebox{\linewidth}{!}{%
\begin{tabular}{lccccccccccccc}
\toprule                                                                                                   
                        &&     & PSNR &     &&     &\clfp &     &&     & t' &     \\
\multicolumn{1}{l}{}    && Circle & PII  & Feature && Cirle & PII & Feature && Cirle & PII & Feature \\
\cmidrule(lr){3-5}\cmidrule(lr){7-9}\cmidrule(lr){11-13}
\cxr &&$24.76$ & $25.06$ & $24.54$ &&$ 0.00$  & $0.00$  & $0.00$  && $0.14$ & $0.25$ & $0.43$ \\ 
\bci &&$46.41$ & $46.48$ & $46.01$ &&$ 0.00$  & $0.00$  & $0.00$  && $0.10$ & $0.26$ & $0.25$ \\  
\odir &&$46.93$ & $46.18$ & $47.14$ &&$ 51.33$ & $61.67$ & $39.33$ && $0.76$ & $0.77$ & $0.78$ \\ 
\bottomrule
\label{tab:fingerprint_types}
\end{tabular}%
}
\end{table}

\paragraph{Experimental details}
We evaluate three types of synthetic anatomical fingerprints (SAFs) to assess whether our findings depend on the specific form of the injected feature.
\emph{Circular fingerprints} are purely synthetic and consist of a gray disk with a fixed radius of 72 pixels, placed at a predefined location within the region typically occupied by image content. 
These fingerprints are visually simple and trivial to detect, which makes automated detection reliable and serves as a controlled baseline.
To study visually more realistic fingerprints, we construct \emph{PII-based fingerprints} using PII, which is known to introduce highly realistic local features and is commonly used in anomaly detection settings~\citep{tan2021detecting}. 
For ChestX-ray14, we select source images containing a medical support device and inpaint the corresponding region into target images at locations where such devices are not present. 
This ensures that no real training sample contains a visually similar feature at the target location.
We experiment with \emph{feature fingerprints derived from image-level labels}. 
Specifically, we use sex for ChestX-ray14, staining type for BCI, and the physical side of the eye for ODIR-2019. 
Since these attributes are real but globally defined, we synthetically localize them by applying PII to all images with the opposite label. 
This allows us to reuse the same classifier used for PII-based fingerprints to detect these features. 
Effectively, this procedure turns a real attribute into a synthetic, localized fingerprint while preserving a robust decision boundary for detection.
To quantify memorization behavior across fingerprint types, we additionally compute the peak signal-to-noise ratio (PSNR). 
Specifically, we compute the PSNR between 1{,}000 generated samples and all images in the training set, and report the maximum value. 
High PSNR values indicate strong pixel-level similarity to a training image and are therefore indicative of potential memorization.

\paragraph{Results}
Results are reported in Tab.~\ref{tab:fingerprint_types} and averaged over three independent runs.
Across ChestX-ray14 and BCI, none of the models reproduce the injected fingerprints, independent of fingerprint type, indicating systematic fairness issues.
The three fingerprint types yield similar trends in both $\mathrm{Cf}^+$ and $t'$.
Circular fingerprints are most easily forgotten (lowest $t'$), while PII-based and feature-based fingerprints produce comparable $t'$ values, suggesting that visually realistic interventions behave more similarly to real attributes.
Overall, the type of fingerprint has only a minor influence on the observed behavior, indicating that our findings are robust to the specific choice of fingerprint construction.

\subsubsection{Impact of Number of Synthetic Anatomic Fingerprint}
\begin{figure*}
    \centering
    \includegraphics[width=\linewidth]{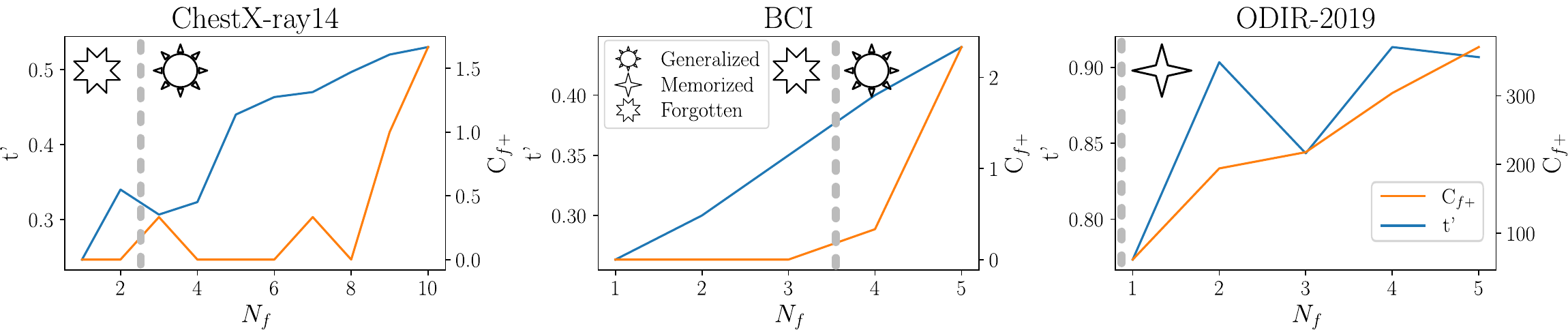}
    \caption{Number of detected fingerprints in the synthetic dataset and memorization indicator $t’$ as a function of the number of fingerprinted training samples $N_f$. The grey line marks the threshold at which models begin reproducing fingerprints. Left to this threshold, fingerprints are either forgotten or memorized together with their training identity; right to it, early signs of generalization emerge for ChestX-ray14 and BCI datasets.}
    \label{fig:multiple_fingerprints}
\end{figure*}

\paragraph{Experimental Details}
We next study how many fingerprinted training samples are required for a diffusion model to generalize a unique feature rather than memorizing or discarding it. 
For each dataset, we vary the number of fingerprinted examples $N_f$ while keeping the remaining training conditions fixed. The objective is to determine the smallest value of $N_f$ for which the model begins to synthesize the fingerprint rather than reproducing exact copies or failing to express it altogether. We visualize the results in \Cref{fig:multiple_fingerprints}.

\paragraph{Results} 
Across all datasets, successful generalization requires a minimum number of fingerprinted examples. For \cxr, early signs of generalization appear at roughly three fingerprinted images, though reproduction remains sparse (one out of 50k generated samples at $N_f=9$).
\bci reaches generalization after approximately four fingerprinted samples, suggesting that simpler datasets require fewer examples.
In contrast, \odir\ shows no evidence of generalization; its limited size causes the model to rely on memorization regardless of $N_f$.
Too few fingerprinted samples lead either to memorization or complete omission; a sufficient number allows the model to form a stable representation supporting controlled synthesis.

\subsubsection{Evaluation Summary}
Across all experiments, we evaluated how dataset size, training duration, model capacity, fingerprint type, and the number of fingerprinted samples influence memorization and generalization in diffusion models. 
Small datasets consistently caused the models to reproduce training images, constituting clear privacy violations. Larger datasets shifted behavior toward broader synthesis but still showed fairness issues when long tail features were underrepresented. 
Varying the training length revealed that extended optimization often led to mode collapse rather than increased memorization, except for very small datasets where memorization persisted. 
Reducing model size lowered the likelihood of reproducing training samples but degraded generative quality and caused the model to forget rare features. 
The type of fingerprint used for the analysis has minimal influence.
We found that a minimum number of fingerprinted examples is required for a model to generalize a unique feature instead of memorizing or omitting it. 
Consequently, diffusion models face a consistent trade-off: limited data or capacity leads to memorization, while attempts to reduce memorization risk often suppress rare but important clinical information.

\subsection{Conditional Image Generation}
In \Cref{sec:unconditional} we showed that unless the SAFs are sufficiently often present in the training dataset, the generated data will either exhibit fairness issues in terms of forgetting long-tail data, or privacy related issues, by memorizing training data. 
However, the experiments were limited to a single family of diffusion model and unconditional image generation. 
Here we extend  to conditional image generation.
The model might be able to quickly pick up rare signals if it is causally connected to the input conditioning. 
One example is given by~\cite{carlini2023extracting}, where an uncommon name, as a unique key in the training dataset, was used to extract training data of text-conditioned diffusion models. 
This type of memorization was possible for training data of Stable Diffusion~\citep{rombach2022high}, which was trained on LAION-400M, a dataset with 400 million text to image pairs~\citep{schuhmann2021laion400mopendatasetclipfiltered}.
We hypothesize there might be a connection between the rarity of the input conditioning and the ability of the diffusion model to memorize. 
To test this, we borrow concepts from information theory, specifically entropy and surprisal, as introduced in \ref{sec:methdo_class_cond}. 
We measure the unpredictability of the input conditioning in terms of entropy, and combine the synthetic fingerprints with the most suprising and least suprising elements of the dataset. 
Unsurprising images in that sense are images, with masks close to the mean masks, or with short and often reccuring text laebls. 
More suprising images, on the other hand, look visually different. The come from rare diseases, misaligned lungs, or very long and specific prompts.

\paragraph{Setup}
To compare high and low surprisal runs, we perform two runs for each modality, where we attach the $N_f$ fingerprints to the most surprising samples in one run and to the least surprising samples in the other run.
For example, under class conditioning, this corresponds to modifying the rarest class and the most common class.
We then generate as many samples as required such that the expected number of fingerprint occurrences in the synthetic data is ten.
We record both the total number of generated samples and the number of samples containing memorized fingerprints.
In \Cref{fig:surprisal}, we show representative examples of the least and most surprising conditionings.

For \emph{class conditioning}, we employ EDM-2 autoguidance, building on the EDM-2 framework~\citep{karras_analyzing_2024,karras2024guiding}.
For \emph{text conditioning}, our setup follows~\cite{rombach2022high}, with the fine-tuning procedure adapted from~\cite{moroianu2025improvingperformancerobustnessfairness}.
As text conditioning, we use the impression sections of the reports.
Due to internal limitations of the language encoder, prompts are truncated after 77 tokens.
Selecting high surprisal images without manual supervision resulted in outliers being chosen.
These included side views, completely white images, or images with excessively large margins.
We therefore manually removed such samples during preprocessing.
For \emph{mask conditioning}, we use ControlNet~\citep{zhang_adding_2023}.
The model is trained according to the recommendations of the original publication for a total of 10{,}000 optimization steps.
We use the finetuned text-conditional model as the initial checkpoint and empty strings as text conditioning.
\emph{Feature conditioning} uses image features extracted with an Inception network following~\cite{dombrowski_image_2024}.
While these features are not medically meaningful, their visual alignment facilitates the generation of realistic images~\citep{dombrowski2025enabling}.
We compute the features on images without SAFs but train the model on images containing SAFs.

For the SAFs, we experiment with four distinct fingerprints in a single run, where we vary the number of occurrences of each fingerprint $|N_f|$.
All of them are inpainted using PII. 
Results can be seen in \Cref{fig:saf_fingerprint_example}.
One fingerprint is unique, while the others appear 5, 10, and 20 times in the training dataset.
Each fingerprint is consistently linked to the conditioning key, as discussed in the respective paragraphs in \Cref{sec:methdo_class_cond}.
For each fingerprint, we train a separate classifier $C_{f}$ to detect fingerprints in the generated dataset.
For training the \cid\ model, instead of training multiple models for each fingerprinted image, we leverage LCMem, a re-identification foundation model~\citep{dombrowski2025lcmemuniversalmodelrobust}.
It takes two images as input and predicts whether they depict the same subject (output 1) or different subjects (output 0).
We fine-tune LCMem for twenty epochs on MIMIC-CXR and apply fingerprint inpainting as a random training transformation.
This encourages the identification model to avoid assigning identity based on the fingerprints and instead focus on other image details.
This behavior is crucial for our disentangled analysis of identity and feature reproduction.
The performance of this model on the balanced real test dataset reached more than 96\% recall.

\paragraph{Dataset}
All models are trained and evaluated on the same dataset and data split, using a single conditioning signal per image.
Specifically, we use a curated subset of MIMIC-CXR restricted to samples with a single positive label to enable class conditional generation.
All images are associated with an available impression and are limited to posterior anterior (PA) views only.
Lung segmentation masks are available for all images.
For feature based generation, we extract feature vectors using an InceptionV3 model pretrained on ImageNet.
The resulting dataset contains 64{,}198 images, split into 38{,}493 training samples, 6{,}278 validation samples, and 19{,}427 test samples, with four different conditioning signals per image.

\paragraph{Metrics}
To evaluate image quality, we use the FID.
For privacy and fairness auditing, we measure the number of detected fingerprints in a dataset of synthetic images, denoted by \clfp, as well as the set of memorized samples quantified by $|q|$, as introduced in \Cref{sec:app_training_details}.
Since we rely on a re-identification model rather than a one vs all classifier for identity detection, the decision rule of \cid\ changes from $C_{id}(x)=1$ to $\max(C_{id}(x, x_{\text{saf}}))$, where $x_{\text{saf}}$ denotes the set of all fingerprint inpainted training images.

\subsubsection{Conditional Image Quality Results}
\begin{figure}
    \centering
    \includegraphics[width=\linewidth]{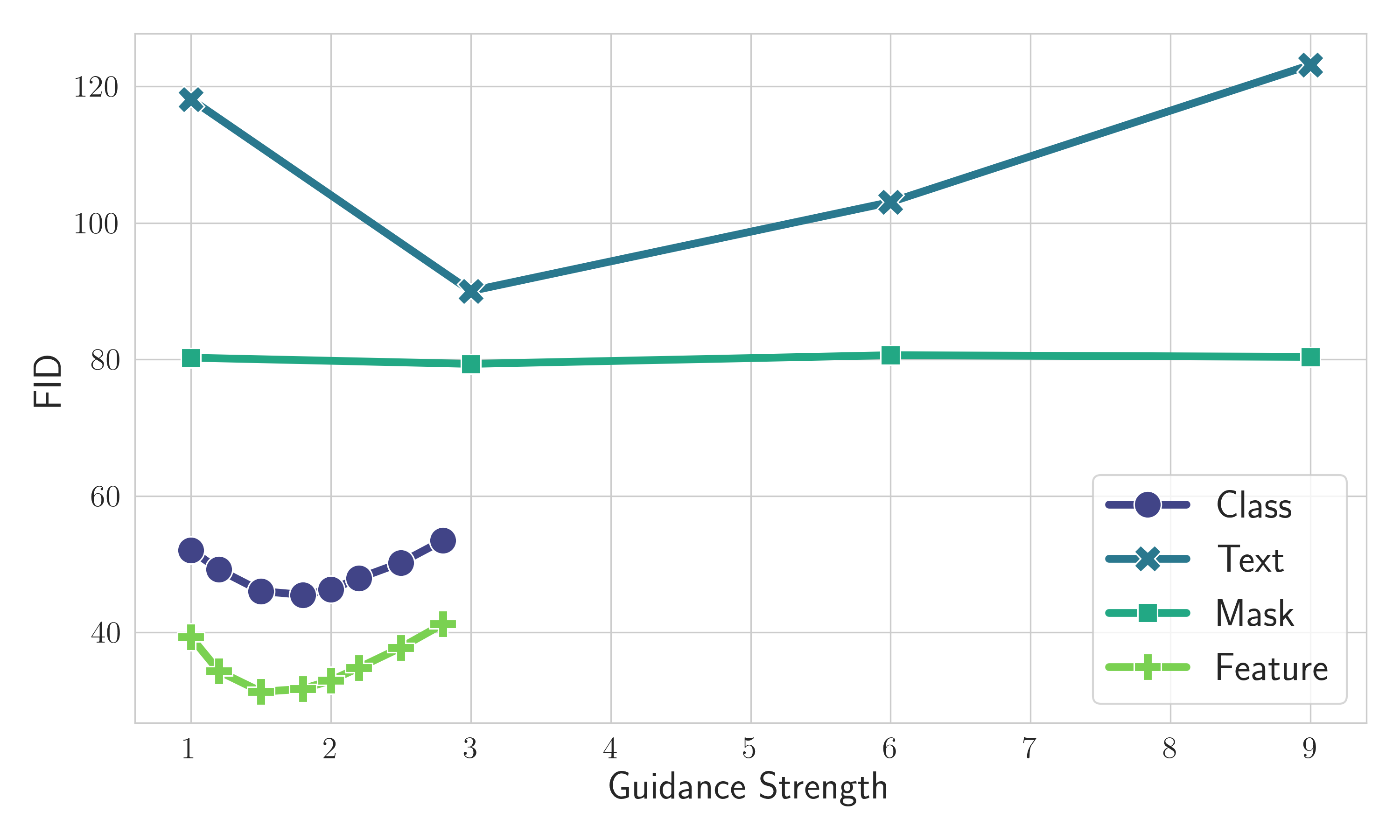}
   \caption{FID as a function of guidance strength for all four conditioning methods. Each curve represents one method trained on MIMIC-CXR. For Stable Diffusion-based architectures (text, mask), guidance strengths range from 0 to 9; for EDM-based architectures (class, feature), from 1 to 3. Lower FID indicates better perceptual quality.}
    \label{fig:fid_vs_guidance_conditional}
\end{figure}
All models are trained until the FID no longer improves.
After convergence, we select the guidance hyperparameter that yields the best FID and compare the overall image quality across all conditioning methods.
For each method, we generate 4{,}000 images sampled from the training distribution’s conditions and compute the FID with respect to the entire test set.
The results are shown in \Cref{fig:fid_vs_guidance_conditional}.
The comparison between text and mask conditional generation is fair, as the only architectural difference arises from the ControlNet extension, including zero convolution layers and additional fine tuning steps for mask conditioning.
Similarly, class conditional and feature conditional generation are directly comparable, as both share the same backbone architecture and differ only in the conditioning head.
While differences between these two groups of methods may partly stem from architectural choices, EDM based architectures generally achieve better image quality than Stable Diffusion based architectures.
Within both groups, conditioning signals with higher information content consistently lead to improved FID scores.
Guidance strength has a notable effect for all methods except ControlNet.
In this case, the model is trained with empty text prompts, such that the unconditional and conditional predictions used for guidance are identical at sampling time.
We therefore fix the guidance scale to one for mask conditional generation.
For all subsequent privacy and fairness experiments, we use the best performing model configuration for each conditioning method.

\subsubsection{SAF Results}
\paragraph{Experimental Details}
To account for the fact that some conditioning signals contain proportionally fewer fingerprints than others, we vary the number of generated samples across experiments.
In class conditional image generation, for example, the rare condition appears 114 times in the training dataset.
Out of these, 36 images are augmented with a fingerprint.
If the model perfectly memorized the training data, we would therefore expect the fingerprint to be generated with probability $36/114$.
To obtain an expected count of 360 fingerprints, we sample the model 1{,}140 times.
For the low surprisal class condition, achieving the same expected number of fingerprints requires sampling 132{,}864 images.
For text, mask, and feature conditioning, this adjustment is simpler, as the conditioning signal is unique for each image.
The only exception is the low surprisal text condition, which corresponds to an impression that appears frequently in the training dataset.
In this case, the impression occurs 50 times in the training data, with 36 of these images augmented with a fingerprint.
Accordingly, we sample 500 images to obtain a comparable expected fingerprint count.
The resulting set of synthetic images should now contain an equal amount of fingerprint across all conditions.
After running the fingerprint classifier \clf, we manually remove clear false positives.
These arise because image generation occasionally fails, producing degenerate samples that can act as adversarial inputs to the classifier.

\paragraph{Results}
\begin{table}[t]
\centering
\resizebox{\linewidth}{!}{%
\begin{tabular}{llrrrr}
\toprule
\multirow{2}{*}{Conditioning} & \multirow{2}{*}{Split} & \multicolumn{4}{c}{$N_f$} \\
\cmidrule(lr){3-6}
& & 1 & 5 & 10 & 20 \\
\midrule
\multirow{2}{*}{Class}
& high & 0 & 0 & 1 -- 1 & 38 -- 22 \\
& low  & 0 & 0 & 0 & 1 -- 1  \\
\midrule
\multirow{2}{*}{Text}
& high & 0 & 55 -- 47  & 111 -- 108 & 195 -- 180 \\
& low  & 0 & 0   & 0   & 0   \\
\midrule
\multirow{2}{*}{Mask}
& high & 0 & 0 & 0 & 0 \\
& low  & 0 & 0 & 0 & 0 \\
\midrule
\multirow{2}{*}{Feature}
& high & 0 & 23 -- 23 & 34 -- 33 & 57 -- 57 \\
& low  & 0 & 0  & 0  & 21 -- 15 \\
\midrule
Expectation & & 10 & 50 & 100 & 200 \\
\bottomrule
\end{tabular}
}%
\caption{Counts of detected fingerprints and memorized fingerprints (\clfp\ -- $|q|$).
Columns indicate the number of fingerprints $N_f$ present in the training dataset.
Experiments are grouped by conditioning type and by whether fingerprints are embedded in low surprisal or high surprisal conditionings.
For truly unique fingerprints ($N_f = 1$), all methods forget them.
In general, fingerprints embedded in low surprisal conditionings tend to be forgotten, while fingerprints embedded in high surprisal images are memorized.
}
\label{tab:fp_pospred_summary}
\end{table}

The results of the class conditional experiments are shown in \Cref{tab:fp_pospred_summary}.
The number of reproduced SAFs differs substantially between the high and low surprisal settings.
SAFs embedded in images with high surprisal conditionings are frequently reproduced.
In contrast, SAFs embedded in low surprisal regions are almost entirely forgotten.
Indeed, only a single fingerprint, corresponding to one class conditional synthetic image, was generated in a low surprisal region.

For mask conditioning, we do not observe a single memorized fingerprint.
While this is a positive outcome for low surprisal regions, the results for high surprisal regions are dominated by a different failure mode.
In these cases, ControlNet fails to generate images conditioned on masks that deviate strongly from the typical mask structure shown in \Cref{fig:Lungentropy}.
A meaningful analysis of mask conditioning for high surprisal outliers would therefore require more advanced methods for robust mask conditional generation.
For low surprisal regions, however, the generated image quality is higher and the masks correctly follow the conditioning signal.
Despite this, even the most frequently occurring fingerprints are not reproduced.
This is likely because the fingerprints are not part of the mask itself and are therefore not directly encoded in the conditioning signal.

\begin{figure}
    \centering
    \includegraphics[width=\linewidth]{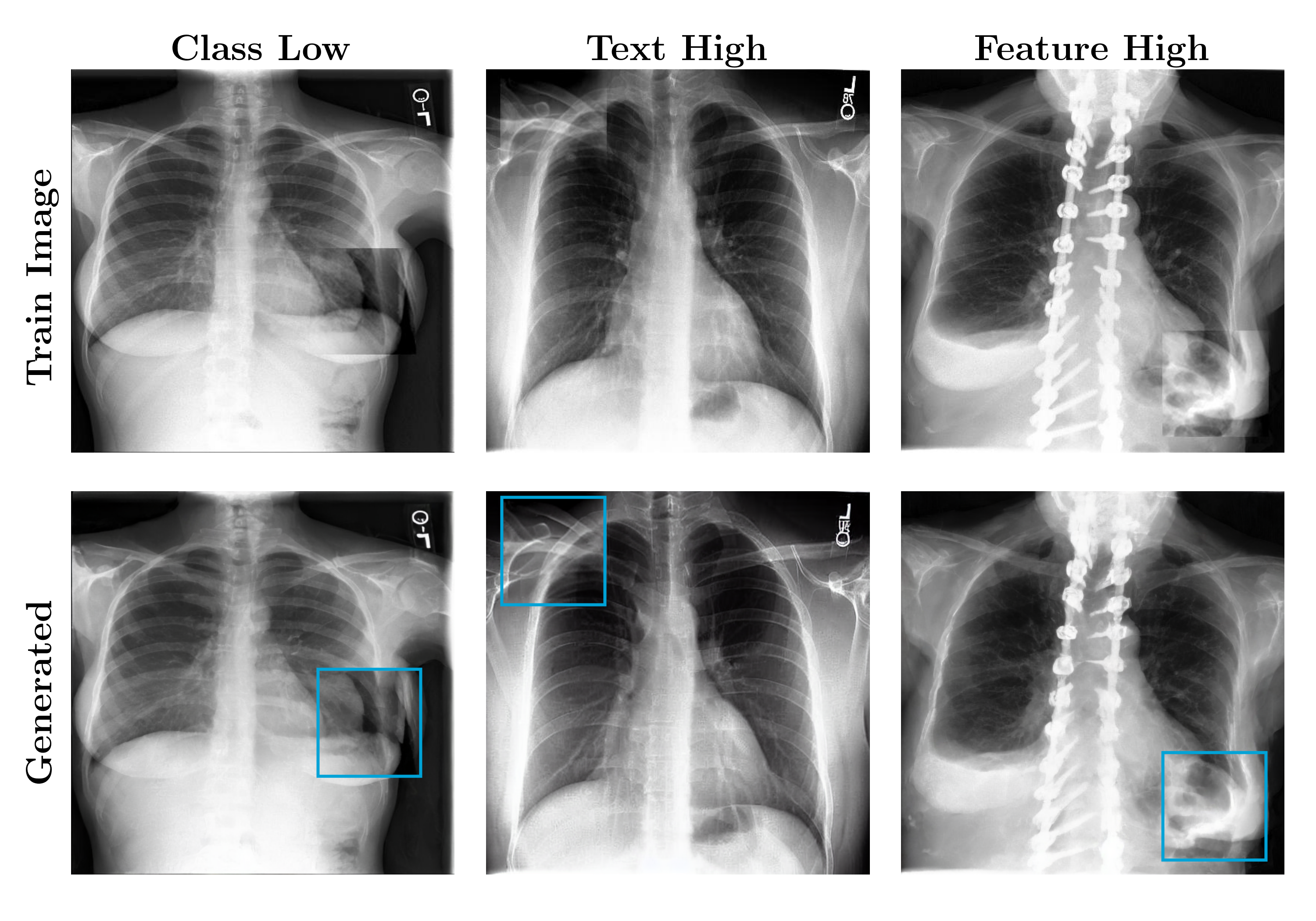}
    \caption{Comparison of training images (top) containing SAFs with generated images that received positive SAF and identity predictions (bottom). The SAF is indicated by the blue box. In all shown cases, the entire training image is reproduced, confirming memorization rather than generalization of the fingerprint alone.}
    \label{fig:memorization_examples}
\end{figure}

Text and feature conditioning produce the largest number of reproduced fingerprints.
Both follow a similar trend in that truly unique fingerprints are not reproduced.
However, once a fingerprint appears multiple times throughout the dataset, models quickly learn to pick up the signal.
For text conditional generation in high surprisal regions, the number of reproduced fingerprints is close to the expected value.
More importantly, almost all images containing a fingerprint are exact memorized reproductions.
This indicates that the model does not generalize well in this regime.
At the same time, the number of memorized samples closely matches the empirical fingerprint distribution in the training data.
While this suggests that additional calibration is required to reduce memorization, it also indicates that the model follows the conditioning signal faithfully and generates a representative synthetic distribution, even for long-tail data.
For feature conditional diffusion models, the number of memorized images is high but lower than the expected value.
Notably, this is the only method that also memorizes fingerprints in low surprisal regions.
Across all methods, the likelihood of reproducing a fingerprint increases with the number of occurrences $N_f$ in the training dataset.
We visualize a selection of memorized samples in \Cref{fig:memorization_examples}.

\subsubsection{Discussion}
From a fairness and long-tail preservation perspective, the results are unambiguous.
Truly unique fingerprints are consistently eliminated, even when encoded through strong conditioning mechanisms such as pseudo-conditional or text-conditional generation.
This indicates that the generative model capacity is insufficient to reliably reproduce such unique signals.

The most interesting comparison is between high and low surprisal text conditional generation.
For $N_f = 20$, the high surprisal text conditional model reproduces the fingerprint multiple times.
This behavior is analogous to a rare name acting as a key to an individual, although the model must have observed the corresponding image several times.
In contrast, embedding the same fingerprint into a frequently occurring, low entropy impression leads to the fingerprint being forgotten.
This is particularly striking because the complete impression appears 50 times in the dataset, with 36 instances inpainted and 20 of them containing the exact same fingerprint.
Yet, out of 500 generated samples, not a single image reproduces the fingerprint.
This behavior can be explained by how the model processes the conditioning signal.
Although the full impression is unique as a sequence, it is decomposed into individual tokens that are each very common.
As a result, the model does not associate the fingerprint with the impression as a whole.
This finding has important implications for diversity.
While reduced diversity has previously been attributed primarily to long-tail effects, our results indicate that head data is also affected.
Uniqueness within frequent classes is diminished, rare artifacts are suppressed, and generated images are pushed closer to the mean.
Consequently, synthetic data becomes increasingly homogeneous and less distinctive.

More generally, high surprisal conditions require careful assessment to ensure that the conditioning mechanism remains effective and that image quality does not degrade.
Among all methods, the class conditional model operating on high surprisal conditions comes closest to genuine generalization.
Approximately one third of the generated images containing SAFs yield negative \cid\ predictions.
However, these images are visually blurry, indicating that the class conditional model struggles to generate high quality samples for these rare classes.

\paragraph{Mitigation Strategies}
From a fairness perspective, clear and targeted evaluation of rare subclasses is essential when working with synthetic data.
Not all real world fingerprints need to be preserved.
However, determining which features can be safely forgotten requires careful analysis, \emph{e.g.}, by measuring downstream task performance on long-tail classes.

From a privacy perspective, truly unique fingerprints are not reproduced under conditional image generation, even for strong conditioning schemes such as pseudo conditional generation.
This is a positive result, as it suggests that if a visual fingerprint deviates sufficiently from the expected image structure, the model learns not to reproduce it.
Under these conditions, conditional generation appears robust against extractable memorization attacks such as those shown in \Cref{fig:realworldcxr}.
Once a fingerprint becomes more common, however, the model begins to reproduce it across conditional, text conditional, and pseudo conditional settings.
This highlights the need for targeted mitigation as rare features transition from unique outliers toward more typical patterns in the training data.

\paragraph{Practical recommendations.}
Our findings translate into concrete guidance for practitioners:
(1)~Avoid high-surprisal conditioning for sensitive data. For example, rare patient names or unique report phrases in text-conditional models can act as retrieval keys that amplify memorization.
(2)~Prefer low-surprisal, structured conditioning (\emph{e.g.}, class labels or segmentation masks), which showed near-zero SAF reproduction.
(3)~Even when memorization is avoided, rare features are systematically suppressed; practitioners generating synthetic training data should verify that clinically relevant rare features are represented.

\section{Limitations}
Our analysis relies on SAFs as controlled proxies for rare or sensitive features. 
While SAFs enable systematic and observable memorization experiments, they do not capture the full semantic complexity of real protected attributes. 
Consequently, memorization behavior observed for SAFs may differ from memorization of naturally occurring patient identifiers or clinically meaningful rare patterns.
Additionally, memorization assessment is sensitive to numerous design choices, including model architecture, training duration, optimization objectives, dataset composition, and preprocessing strategies, which can lead to inconsistencies across evaluations.
We remain consistent within our desing choices, but results may vary for other approaches.
In the conditional setting, the compared architectures differ not only in conditioning type but also in backbone design (\emph{e.g.}, class-conditional DDPM vs.\ text-conditional Stable Diffusion vs.\ ControlNet), making it difficult to attribute observed differences to conditioning alone.
Nevertheless, the consistent relationship between conditioning surprisal and memorization behavior across all tested architectures suggests that the information-theoretic perspective captures a fundamental property of the conditioning mechanism.
Furthermore, the proposed memorization indicator provides evidence about a model's capacity to memorize training samples, but it cannot establish the absence of memorization. 
In particular, failure to detect memorization should not be interpreted as a hard privacy guarantee, especially for large or highly redundant datasets.
Additionally, some cases classified as forgetting may instead represent poor-quality generalization that falls below the classifier detection threshold.
Whether such low-fidelity reproductions are useful for downstream clinical tasks, or whether they degrade performance similarly to complete omission, remains an open question that requires task-specific evaluation.
Finally, our notion of fairness is limited to representational and distributional fairness, whether rare features are preserved in generated outputs, and does not address clinical downstream fairness such as demographic parity or equalized subgroup performance.

\section{Conclusion}
In this work, we introduced a framework to analyze the interplay between fairness and privacy in diffusion based image generation.
We proposed synthetic anatomical fingerprints as a controlled data-intervention to study whether diffusion models generalize, memorize, or forget rare and unique features present in their training data.
Across both unconditional and conditional training setups, we observe a consistent behavior: unique features are either memorized or entirely forgotten, with little evidence of meaningful generalization.
For conditional models, we identify a key dependence on the surprisal of the conditioning signal.
When unique attributes are embedded in images associated with low surprisal conditions, they tend to be forgotten.
In contrast, embedding the same attributes under high surprisal conditions leads the model to memorize not only the feature itself but the entire image.
These observations directly inform mitigation strategies for both fairness and privacy.
From a fairness perspective, evaluation should explicitly focus on long-tail performance, as aggregate metrics may hide systematic failures on rare subclasses.
Not all real world fingerprints need to be preserved, but determining which features can be safely forgotten requires careful evaluation, for example through downstream task performance on long-tail classes.
From a privacy perspective, our results indicate that truly unique visual fingerprints are not reproduced under unconditional or conditional image generation, even for strong conditioning schemes such as pseudo-conditional training.
This suggests that, as long as a fingerprint is sufficiently atypical relative to the expected image structure, conditional generation remains robust against extractable memorization attacks as illustrated in \Cref{fig:realworldcxr}.
However, once such fingerprints become more frequent and visually plausible, models begin to reproduce them across conditional, text-conditional, and pseudo conditional settings, highlighting the need for targeted mitigation when rare features move from the tail toward the head of the data distribution.


\acks{HPC resources were provided by the Erlangen National High Performance Computing Center (NHR@FAU), under the NHR projects b143dc and b180dc. NHR is funded by federal and Bavarian state authorities, and NHR@FAU hardware is partially funded by the DFG - 440719683. We acknowledge the use of Isambard-AI National AI Research Resource (AIRR)~\citep{mcintosh2024isambard}. Isambard-AI is operated by the University of Bristol and is funded by the UK Government's DSIT via UKRI; and the Science and Technology Facilities Council [ST/AIRR/I-A-I/1023]. The authors received funding from the ERC-project MIA-NORMAL 101083647, DFG 513220538, 512819079, and by the state of Bavaria (HTA).}

%
\ethics{The work follows appropriate ethical standards in conducting research and writing the manuscript, following all applicable laws and regulations regarding treatment of animals or human subjects.}

\coi{MD declares no conflicts of interest. BK is a consultant for ThinkSono Ltd. and a co founder of Fraiya Ltd. Neither company was involved in the conception, design, execution, analysis, or interpretation of the work presented here. }

\data{All datasets used in this study are publicly available. Each dataset is referenced at the appropriate locations in the manuscript, together with links or citations that allow readers to access the data directly.}

\bibliography{main}

%
\clearpage
\appendix
\section{Derivation of t'}
\label{sec:estimation_method}
\begin{figure}
      \centering
      \fbox{
      \includegraphics[width=0.95\linewidth]{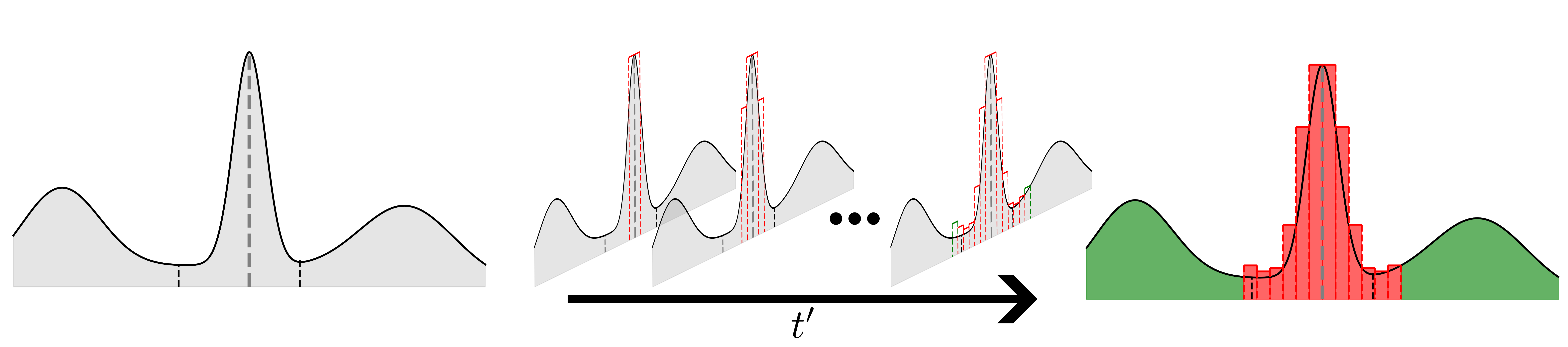}
      }
    \caption{Illustration of our estimation method in 1D. The grey line denotes the query image $\bfxp$. The estimation method iteratively increases the search space in the latent space of the generative model. The green area corresponds to image regions resulting in non-privacy concerning generated samples, while the red area is considered critical. }
    \label{fig:illustrationofmethodin1D}
\end{figure}

\begin{figure}
	\centering
	\fbox{
		\includegraphics[width=0.95\linewidth]{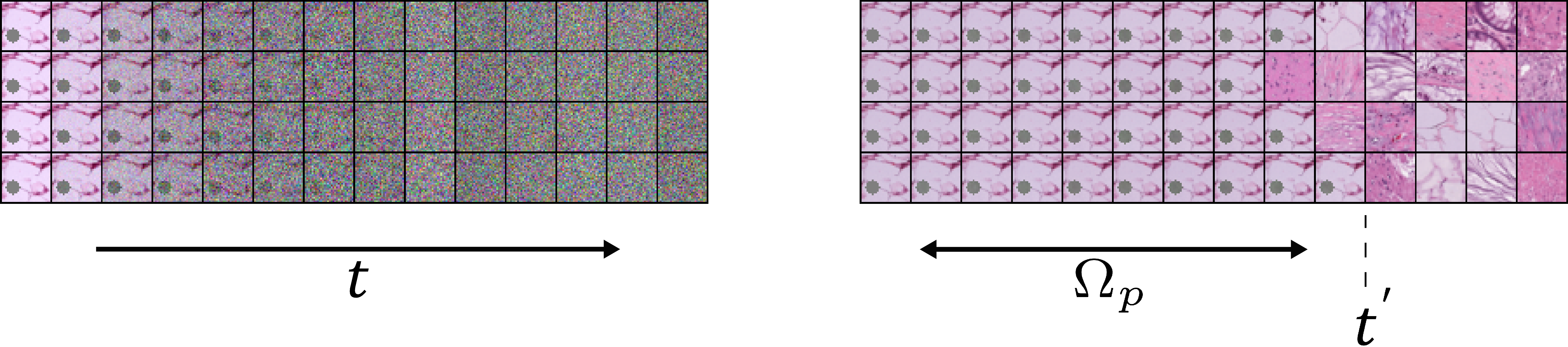}
	}
	\caption{Illustration of the reverse diffusion process. Left shows query images  $\bfx_{t,p}$ for $t \in \left[0, 0.7\right]$. Right shows the resulting sample.}
	\label{fig:reversediffusion}
\end{figure}

\citet{song2020score} show that the reverse diffusion process of the SDE can be modeled as a deterministic process as the marginal probabilities can be modeled deterministically in terms of the score function. As a result, the problem of learning transition kernels simplifies to an ODE: 
\begin{align}
    \ud \bfx = \Big[\bff(\bfx, t) - \frac{1}{2} g(t)^2\nabla_\bfx \log p_t(\bfx)\Big] \ud t, \label{eqn:deterministic_flow}
\end{align}
Solving Eqn. \ref{eqn:deterministic_flow} enables exact likelihood computation. 
However, this does not account for the fact that images in the immediate neighborhood, like slightly noisy versions of $\bfxp$, are not anonymous. 
Consequently, we are interested in computing $q(p)$, which is defined as the likelihood of reproducing any sample within $\Omega_p$, which is the region of the image space that is similar enough to $\bfxp$ that it raises privacy concerns: 
\begin{align}
   q(p) = \int_{\Omega_p} p_s(\bfx)\ud\bfx \label{eq:pofp}.
\end{align}
We determine this region by training a classifier tasked with detecting whether the image belongs to the image class \clf.
To search through the image manifold, we make use of the reverse diffusion process centered around the SAF image $\bfxp$ defined as $p_{t,b}  \coloneqq p(\bfx_t \mid \bfxp) =  \mathcal{N}(\tilde{\bfx}; \bfxp, \sigma_t^2 \bfI)$ for $\bfx(s)$ to $\bfx(t)$, where $0 \leq t \leq T$. 
We can employ the diffusion process centered around this image to sample from the neighborhood and then use the learned reverse diffusion process to generate noisy samples $\bfx_{t,p}$. Then we can use this as starting image for the reverse diffusion process to sample $\bfx_{t,p}'$:
\begin{multline}
   q(p) = \int_{\Omega_{p}} p_s(\bfx)\ud\bfx \approx \int_0^{t'} p_s(\bfx_{t,p}) \ud\bft \\
   = \int_0^{t'}\mbb{E}_{p(\bfx_{t,p})}\big[    p(\bfx'_{t,p})\big]\ud\bft .
\end{multline}
Technically, we could employ exact likelihood computation to estimate $q(p)$ but this would require integrating over the continuous image-conditioned diffusion process, which would be intractable in practice. Therefore, we propose to approach and estimate this integral by computing the Riemann sum of this integral and give an upper bound estimate for it using the upper Darboux sum: 

\begin{multline}
\int_0^{t'}\mbb{E}_{p(\bfx_{t,p})}\big[    p(\bfx'_{t,p})\big]\ud\bft = \\
\sum_{t} (\sigma_t - \sigma_{t-1}) \mbb{E}_{p(\bfx_{t,p})}\big[p(\bfx'_{t,p})\big] \\
\leq \sum_{i=0}^{t'} \sup_{t \in\left[t_i, t_{i+1}\right]} (\sigma_{t_{i+1}} - \sigma_{{t_i}})\mbb{E}_{p(\bfx_{t,p})}\big[p(\bfx'_{t,p})\big], \label{eq:full_equation_estimate} 
\end{multline}
which approaches the real value for steps that are small enough. We can compute this value by using $\bfxp$ as a query image and estimating the expectation by performing Monte-Carlo sampling but this would be computationally infeasible due to the complexity of exact likelihood estimation.
We sketch this 1D search in \Cref{fig:illustrationofmethodin1D} and visually in \Cref{fig:reversediffusion}.

\paragraph{Estimation Algorithm}
\begin{algorithm}[h]
\caption{Upper bound likelihood estimation algorithm}
\label{alg:CKB}
\KwIn{$M$, $s_\theta(\bfx, t)$, $c_{f}(\bfx)$, $c_{ID}(\bfx)$, $\bfxp$}
\KwOut{$t'$}
\For{$t \leftarrow 1$ \KwTo $0$}{
    \For{$m \leftarrow 1$ \KwTo $M$}{
        $\bfx_{t,p} \gets p(\bfx_t \mid \bfx_p)$\;
        \For{$\tilde{t} \leftarrow t$ \KwTo $0$}{
            $\bfx_{t,p}' \gets s_\theta(\bfx_{t,p}', \tilde{t})$\;
        }
        $\bfxp' \gets \bfx_{t,p}'$\;
        \If{$c_{f}(\bfx)$ is \textbf{True} \textbf{and} $c_{ID}(\bfx)$ is \textbf{True}}{
            \Return $t$\;
        }
    }
}
\end{algorithm}
Given $\bfxp$, we define $q_M(p|x_{t,p})$ as the estimate of a sample belonging to $\Omega_p$ for a given diffusion step $t$.
We then define $t' \coloneqq \text{max}(\mathbb{T})$, where $\mathbb{T} \coloneqq \{ \forall t \colon q_M(p|x_{t,p}) > 0 \}$.
The parameter $M$ allows us to trade off accuracy for computation time by choosing the number of generated samples.
In Alg. \ref{alg:CKB} we describe our proposed algorithm to compute the indicator $t'$. 
To do an exhaustive search we set the step size to be the same as the sampling step size, start from the maximum value, and go to the minimum value. 
Since this computation takes too long to be feasible, we experiment with increased step sizes. 
To improve the computation time even further it is straightforward to change the algorithm to a binary search version or to increase the sampling step size.

\paragraph{Practical guidance}
Exhaustive evaluation of $t'$ over all training images is infeasible for large datasets.
We recommend a targeted strategy: compute $t'$ only for samples identified as high-risk by domain experts or by pre-screening with \clf~and \cid.
This keeps the computational cost manageable while focusing auditing effort where it matters most.

\paragraph{Intuition} We model the image space using the learned distribution of the score function $\nabla_{\tilde{\bfx}} \log p_{\sigma_i}(\tilde{\bfx} \mid \bfx)$ by reversing the diffusion process and checking when the model starts to ``break out'' by generating images classified as different samples. 
For large $t$, the learned marginals $p(\bfx, t)$ span the entire image space. Importantly, by definition of the diffusion process, the distribution approaches the same distribution as the sampling distribution of the diffusion process if $\sigma_t$ gets large enough $p_{\sigma_N}(\tilde{\bfx} \mid \bfxp) \sim  \mathcal{N}(\bfx; \textbf{0}, \sigma_N^2\bfI)$. However, for lower $t$ the model has learned that the distribution collapses towards a single training image $\bfxp$. Essentially, it has modeled part of the subspace as a delta distribution around $\bfxp$. 
We want to estimate how far back in the diffusion process we have to go for the model to start to produce different images. 
The boundary $\Omega_p$ is defined as all images that would collapse towards this training image, estimated using the classifiers. 
Fig. \ref{fig:illustrationofmethodin1D} illustrates this process in one dimension. 
The indicator t' is then the strength of the perturbation function according to the definition of discoverable memorization introduced in Sec. \ref{Sec:background}.  
Note that this is different from simply defining a variance that is large enough for the classifiers to fail, as $s_{\theta}(\bfxp, \sigma_t)$ was trained to revert this noise.
Fig. \ref{fig:reversediffusion} illustrates how this looks in image space.

\noindent\textbf{Computational Overhead:}
Our proposed method computes $t'$ through forward passes of the diffusion model, making its computational cost equivalent to that of image sampling. 
The hyperparameter $M$ determines the trade-off between the accuracy of $t'$ and computational overhead, scaling linearly with $M$. For instance, with $M=16$, ensuring privacy for an image requires 16 times the computational cost of generating a single sample.

\section{Model Size}

\begin{table}
        \caption{Model architecture for the unconditional U-Net used as backbone for the diffusion model. The standard value for the number of channels is $c=128$.}
        \centering
        \resizebox{\linewidth}{!}{%
        \begin{tabular}{lccccc}
        & \# Trainable params  & Down blocks &Channels / layer &  Layers / block \\

          \toprule
        \textbf{Default}  & \num{113675524}& 6 & c,c,2c,2c,4c,4c & 2\\
        \midrule
        Model 1           & \num{77364740} & 6 & c,c,2c,2c,4c,4c  &   1\\
        Model 2           & \num{71439108} & 5 & c,c,2c,2c,4c,4c  &   2\\
        Model 3           & \num{49558020} & 5 & c,c,2c,2c,4c,4c  &   1\\
        Model 4           & \num{28484612} & 4 & c,c,2c,2c,4c,4c  &   2\\
        Model 5           & \num{28448388} & 6 & c/2,c/2,c,c,2c,2c&   2\\
        \bottomrule
        \end{tabular}%
        }
    \label{tab:model_size_architecture}
\end{table}

In Tab. \ref{tab:model_size} we summarize the different hyperparameters used to define the backbone architecture of the diffusion model. 

\section{Information Content of Conditioning Modalities}
\label{sec:conditing}

To characterize the strength and diversity of different conditioning signals used for generative modeling, we quantify the information content of each modality in bits.
All conditioning types are treated as random variables with an associated probability model.
The information content of a specific conditioning instance is measured through its surprisal
\begin{equation}
I(x) = -\log_{2} p(x),
\label{eq:surprisal_general}
\end{equation}
while the average information content of a conditioning source is described by its Shannon entropy
\begin{equation}
H(X) = \mathbb{E}[I(X)].
\label{eq:entropy_general}
\end{equation}

This framework allows a direct comparison across conditioning types that differ in dimensionality, representation, or domain.
Each modality defines its own probability model: a multivariate Gaussian for continuous feature vectors,
a multivariate Bernoulli model for token presence,
a pixel-wise Bernoulli field for segmentation masks,
and a categorical distribution for class labels.

\subsection{Gaussian Model for Continuous Feature Conditioning}

Let $z \in \mathbb{R}^{d}$ denote a feature vector extracted from an input image.
We model the distribution of all feature vectors by a multivariate Gaussian
\begin{equation}
z \sim \mathcal{N}(\mu, \Sigma),
\label{eq:gaussian_feature}
\end{equation}
with empirical mean $\mu$ and covariance $\Sigma$.

The differential surprisal of a feature vector is given by
\begin{align}
I_{\text{cont}}(z)
&= -\log p(z) \nonumber\\
&= \tfrac{1}{2}(z - \mu)^{\top}\Sigma^{-1}(z - \mu)
  + \tfrac{1}{2}\log\det(2\pi\Sigma),
\label{eq:surprisal_continuous}
\end{align}
expressed in nats.
We convert this quantity to bits by dividing by $\log 2$.

Differential entropies are scale dependent and may be negative.
To enable comparison with discrete conditioning modalities, we introduce a scalar quantization with step size $\delta$.
The resulting discrete equivalent surprisal is
\begin{equation}
I_{\text{disc}}(z)
= \frac{I_{\text{cont}}(z)}{\log 2}
  + d \log_{2}\!\left(\frac{1}{\delta}\right),
\label{eq:surprisal_discrete}
\end{equation}
and the corresponding entropy estimate is
\begin{equation}
H_{\text{disc}}(Z)
= \frac{1}{N}\sum_{i=1}^{N} I_{\text{disc}}(z_{i}).
\label{eq:entropy_discrete}
\end{equation}

This quantity measures the number of bits required to encode a feature vector at resolution $\delta$.

\subsection{Token Presence Model for Text Conditioning}

Let $T$ denote the set of all tokenizer tokens that appear anywhere in the dataset.
For each text prompt, we record a binary vector $x \in \{0,1\}^{|T|}$ indicating whether each token is present at least once in the prompt.

For each token $t \in T$, its marginal probability of appearing in a prompt is given by
\begin{equation}
p_{t}
= \frac{\text{number of prompts containing } t}
       {\text{total number of prompts}}.
\label{eq:token_marginal}
\end{equation}

For simplicity, we assume that all tokens appear independently of each other.
While this assumption does not hold in practice, modeling the full joint distribution would require substantially more data, which is infeasible for small medical text datasets.
Furthermore, for our relative comparisons, this approximation is sufficient, as we are primarily interested in contrasting low- and high-surprisal text prompts rather than estimating absolute likelihoods.

Under an independent Bernoulli model, the surprisal of a prompt $x$ is
\begin{equation}
I(x)
= \sum_{t \in T}
    \bigl[
        x_{t}\,(-\log_{2} p_{t})
        + (1 - x_{t})\,(-\log_{2}(1 - p_{t}))
    \bigr].
\label{eq:surprisal_text}
\end{equation}

This score quantifies the number of bits required to encode which tokens appear in the prompt, based on their empirical frequencies.
The average text-conditioning entropy is then given by
\begin{equation}
H_{\text{text}}
= \mathbb{E}[I(X)]
= -\sum_{t \in T}
        \bigl[
            p_{t} \log_{2} p_{t}
            + (1 - p_{t}) \log_{2}(1 - p_{t})
        \bigr].
\label{eq:entropy_text_token}
\end{equation}

\subsection{Pixel-Wise Bernoulli Field for Mask Conditioning}

\begin{figure}
    \centering
    \includegraphics[width=0.5\linewidth]{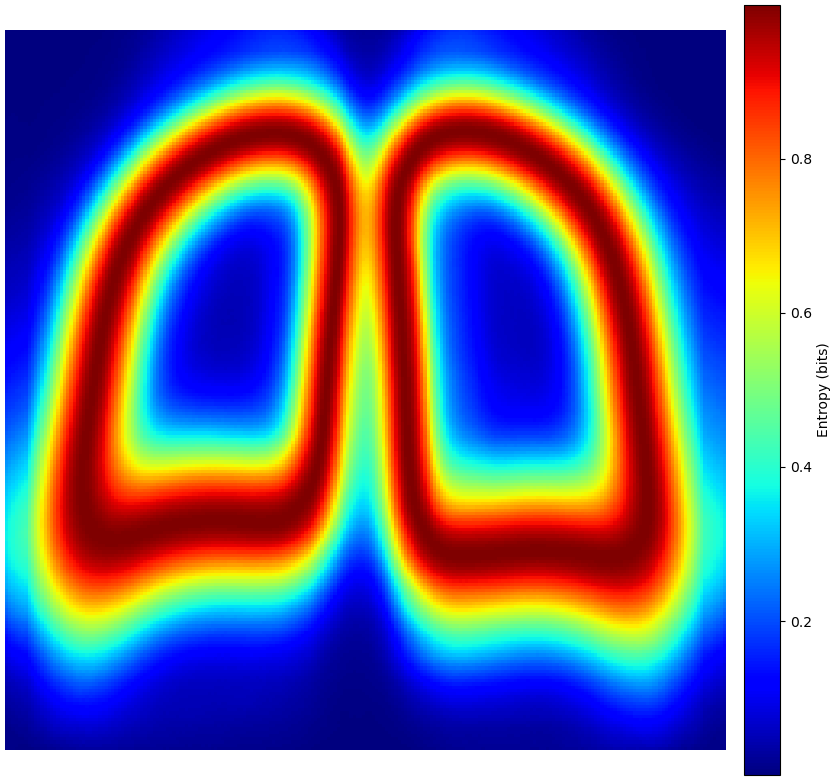}
    \caption{Pixel-wise entropy map of lung segmentation masks.}
    \label{fig:Lungentropy}
\end{figure}

For segmentation-based conditioning, we consider a binary mask
$m \in \{0,1\}^{H \times W}$ and estimate, for each pixel location $(u,v)$,
the empirical probability of being foreground as
\begin{equation}
p_{uv}
= \frac{\text{number of masks where } m_{uv} = 1}
       {\text{number of masks}}.
\label{eq:mask_pixel_prob}
\end{equation}

Each pixel is modeled as an independent Bernoulli random variable.
The surprisal of a specific mask $m$ is then given by
\begin{align}
I(m)
&= \sum_{u,v} m_{uv}\,(-\log_{2} p_{uv}) \nonumber\\
&\quad + \sum_{u,v} (1 - m_{uv})\,(-\log_{2}(1 - p_{uv})).
\label{eq:surprisal_mask}
\end{align}

The pixel-wise entropy map is defined as
\begin{equation}
H_{uv}
= -p_{uv} \log_{2}(p_{uv})
  - (1 - p_{uv}) \log_{2}(1 - p_{uv}),
\label{eq:entropy_pixel_mask}
\end{equation}
and the total mask entropy is given by
\begin{equation}
H_{\text{mask}} = \sum_{u,v} H_{uv}.
\label{eq:entropy_mask_total}
\end{equation}

This quantity represents the number of bits required to encode a segmentation
mask when each pixel is drawn independently from its empirical Bernoulli
distribution.
Visually, the entropy map highlights pixels that are most and least likely
to belong to the lung region, as illustrated in \Cref{fig:Lungentropy}.

\subsection{Categorical Distribution for Class-Conditional Conditioning}

For class-conditioned generation, we consider a discrete random variable
$y \in \{1,\dots,K\}$ with empirical class probabilities
\begin{equation}
p_{k}
= \frac{\text{number of samples with class } k}
       {\text{total number of samples}}.
\label{eq:class_prob}
\end{equation}

The surprisal of observing class $k$ is
\begin{equation}
I(k) = -\log_{2} p_{k},
\label{eq:surprisal_class_cond}
\end{equation}
and the Shannon entropy of the class distribution is
\begin{equation}
H(Y) = -\sum_{k=1}^{K} p_{k}\log_{2} p_{k}.
\label{eq:entropy_class_cond}
\end{equation}

This entropy quantifies the number of bits required to encode class labels
and reflects the inherent uncertainty of the class-conditioned generation task.

\subsection{Unified Interpretation}

All conditioning modalities are now expressed through information content in bits. 
Despite differing statistical structures, each modality quantifies the amount of uncertainty or variability present in the conditioning signal. 
This unified framework allows direct comparison of conditioning strength, complexity, and informativeness across feature, text, mask, and label based conditioning strategies.

\paragraph{Limitations}
The information measures presented here rely on simplifying assumptions and their absolute magnitudes should be interpreted with caution. The text and mask based formulations assume statistical independence across tokens and pixels, although real prompts exhibit syntactic structure and masks contain rich spatial correlations. The feature based estimates assume that the high dimensional latent space is well described by a single multivariate Gaussian, despite the fact that deep representations are often multi modal or heavy tailed. Continuous feature densities must be converted to bits using a chosen quantization scale, which adds an unavoidable offset determined by the selected bin width. In addition, all estimates are subject to finite sample bias. Rare features, rare tokens, and uncommon mask configurations are difficult to estimate reliably and lead to biased entropy values. The preprocessing steps required for each modality, such as resizing masks, truncating text prompts, or relying on a fixed pretrained feature extractor, introduce additional dependencies that influence the probability models.
Although expressing all conditioning signals in bits provides a unified view, comparability across modalities remains imperfect. Bits derived from pixel level Bernoulli variables do not carry the same semantics as bits derived from token presence or quantized latent vectors. Finally, entropy measures the variability of the conditioning source, not its usefulness for generation. 
For these reasons the reported values should be viewed as relative indicators of conditioning complexity rather than precise absolute information quantities.

\end{document}